\documentclass[letterpaper, 10 pt, conference]{ieeeconf}  

\IEEEoverridecommandlockouts                              

\usepackage{amsmath} 
\usepackage{amssymb}  

\usepackage{comment}
\usepackage{makecell} 
\usepackage{multirow} 
\usepackage{graphicx}

\usepackage{pgfplots}
\usepackage{xcolor}
\usetikzlibrary{calc}
\usepackage{xspace}
\makeatletter
\DeclareRobustCommand\onedot{\futurelet\@let@token\@onedot}
\def\@onedot{\ifx\@let@token.\else.\null\fi\xspace}

\def\eg{\emph{e.g}\onedot}

\makeatother

\newcommand{\dualval}[2]{\makebox[2.2em][r]{#1}\textbar\makebox[2.2em][l]{#2}}
\newcounter{showsep}

\title{\LARGE \bf
TOM-GS: Robust Thermal Odometry and Mapping with \\Depth-Guided Gaussian Splatting
}

\author{Tianhao Zhou$^1$, Shengqian Wang$^1$, Zhihao Zhan$^2$,\\
Yuhang Ming$^3$,~\IEEEmembership{Member,~IEEE,} Jianzhu Huai$^{1,*}$,~\IEEEmembership{Member,~IEEE}
\thanks{$^{1}$State Key Lab of Info Engineering in Surveying, Mapping, and Remote Sensing, Wuhan University, 129 Luoyu Road, Wuhan, Hubei 430079, China}%
\thanks{$^2$TopXGun Robotics Limited, Nanjing, Jiangsu 211100, China}%
\thanks{$^3$School of Computer Science, Hangzhou Dianzi University,
Hangzhou, Zhejiang 310018, China}%
\thanks{$^{*}$Corresponding author: jianzhu.huai@whu.edu.cn}
}

\begin{document}

\maketitle
\thispagestyle{empty}
\pagestyle{empty}

\begin{abstract}
Thermal infrared cameras offer robustness to darkness, smoke, and illumination changes, making them attractive for robotic deployment in visually degraded environments. However, most existing thermal SLAM systems are geometric and produce sparse maps, while recent learning-based odometry and neural mapping methods are primarily developed for RGB imagery.

We present \textbf{TOM-GS}, a monocular thermal odometry and mapping system that integrates learning-based dense tracking with depth-guided 3D Gaussian Splatting for dense reconstruction. Monocular depth priors are used to stabilize thermal odometry and supervise Gaussian geometry, improving scale consistency and robustness in low-texture scenes.

Experiments on the RRXIO and VIVID benchmarks show that learned thermal odometry outperforms classical geometric pipelines and that thermal sensing provides improved robustness over RGB under degraded illumination. Furthermore, Gaussian Splatting enables high-quality dense thermal reconstruction in challenging environments.
\end{abstract}

\section{INTRODUCTION}

Thermal infrared cameras operating in the long-wave infrared band (8–14 $\mu$m) provide measurements that are robust to illumination changes, dust, and smoke~\cite{daftryRobustVisionBasedAutonomous2020,khattakKeyframeBasedThermal2020,vanmanenFirebotSLAMThermalSLAM2023}. This property makes them particularly attractive for robotic deployment in visually degraded environments such as subterranean exploration, search and rescue, and firefighting~\cite{nguyenReviewModernThermal2021a, chen4DRadarPRContextaware2025}. As autonomy moves beyond well-lit indoor and outdoor scenes, thermal sensing offers a complementary modality to conventional RGB cameras for simultaneous localization and mapping (SLAM) \cite{huaiObservabilityAnalysisKeyframebased2022}.

Most existing thermal SLAM systems extend geometric pipelines originally designed for RGB imagery. Representative approaches adapt ORB-SLAM~\cite{camposORBSLAM3AccurateOpensource2021} or DSO~\cite{gaoLDSODirectSparse2018} to thermal inputs~\cite{vanmanenFirebotSLAMThermalSLAM2023,wuCatadioptric2024}. While these geometric methods can achieve reasonable tracking accuracy in controlled scenarios, they typically produce sparse maps and remain fragile under low-texture conditions.

Recent advances in learning-based visual odometry (VO), such as DROID-SLAM~\cite{teedDROIDSLAMDeepVisual2021} and DPVO~\cite{teedDeep2023}, have shown strong robustness and dense tracking capability on RGB data. In parallel, neural scene representations, including NeRF~\cite{mildenhallNeRFRepresentingScenes2022} and 3D Gaussian Splatting (GS)~\cite{kerbl3DGaussianSplatting2023}, have enabled high-quality dense mapping and novel-view synthesis. Several GS-based SLAM systems~\cite{matsukiGaussian2024,keethaSplaTAMSplatTrack2024} integrate learned tracking with Gaussian scene representations, achieving impressive performance in RGB-based settings.

Recent advances in monocular depth estimation have further enabled dense geometric priors for monocular reconstruction. Foundation models such as Depth Anything V2~\cite{yangDepthAnythingV22024} and Metric3D~\cite{yinMetric3DZeroshotMetric2023} have shown remarkable generalization across diverse scenes, and have been successfully integrated into RGB-based SLAM and GS systems to improve depth completeness. Although these models can also provide dense depth priors for thermal images, their predictions are generally less reliable.


While monocular depth priors have become an effective source of dense geometric supervision in RGB-based SLAM, their application to monocular thermal SLAM is far from straightforward. The weak textures and distinct radiometric characteristics of thermal imagery lead to unreliable feature correspondences and weak geometric constraints, posing significant challenges for both learning-based odometry and dense reconstruction. Although recent studies have explored thermal depth and pose estimation~\cite{shinMaximizingSelfsupervisionThermal2022}, a unified framework combining learning-based odometry, monocular depth priors, and GS-based mapping remains unexplored.

In this paper, we present TOM-GS, a robust monocular thermal odometry and mapping system that integrates learning-based dense tracking with depth-guided 3D Gaussian Splatting. To improve robustness under low-texture thermal conditions, a learning-based odometry frontend is adopted to establish dense visual correspondences beyond conventional handcrafted features. Furthermore, to compensate for weak geometric cues in thermal imagery, monocular depth priors are used to provide geometric supervision for both pose estimation and Gaussian optimization. By jointly leveraging learned visual correspondences and depth priors, TOM-GS achieves more stable tracking, improved scale consistency, and higher-quality dense reconstruction in challenging thermal environments.

In summary, our main contributions are:
\begin{itemize}
\item \textbf{Thermal odometry and mapping with GS.}  
We present TOM-GS, a learning-based monocular thermal odometry and mapping system that integrates dense tracking with 3D Gaussian Splatting for high-quality thermal scene reconstruction.

\item \textbf{Depth priors for odometry and mapping.}
We incorporate monocular depth priors to enhance learned thermal odometry and to supervise Gaussian geometry during mapping, improving scale consistency and robustness in low-texture thermal environments.

\item \textbf{Extensive evaluation.}  
Through experiments on RRXIO and VIVID, we demonstrate the advantages of learned odometry over classical geometric thermal SLAM, and highlight the robustness of thermal sensing compared to RGB imagery under degraded illumination.

\end{itemize}

\section{RELATED WORK}
\subsection{Thermal Visual Odometry and SLAM}

Thermal cameras have been widely explored for motion estimation and SLAM in visually degraded environments, where visible-spectrum cameras often suffer from illumination variations or complete darkness. Early thermal SLAM systems primarily adapt classical geometric pipelines to thermal imagery, including extensions of ORB-SLAM and DSO-style frameworks~\cite{vanmanenFirebotSLAMThermalSLAM2023,wuCatadioptric2024}. To improve the perceptual quality of thermal imagery, several enhancement and preprocessing strategies have been explored. Representative approaches include self-supervised thermal enhancement methods~\cite{shinMaximizingSelfsupervisionThermal2022}, feature-preserving enhancement techniques such as Fieldscale~\cite{gilFieldscale2024}, and conventional histogram-based methods such as CLAHE. These techniques can improve image contrast and structural visibility, thereby benefiting downstream perception and localization. Nevertheless, most existing thermal SLAM systems still rely on handcrafted feature pipelines and sparse geometric representations, limiting robustness under challenging thermal conditions.

Learning-based visual odometry has recently shown superior robustness compared with traditional feature-based methods. Systems such as DROID-SLAM~\cite{teedDROIDSLAMDeepVisual2021} leverage dense learned correspondences and recurrent optimization to achieve accurate tracking across diverse environments. Such capabilities are particularly beneficial for thermal imagery, where low texture often limits the reliability of handcrafted features. However, the application of learning-based odometry to dense monocular thermal SLAM remains largely unexplored.

\subsection{3D Gaussian Splatting for SLAM}



Splat-SLAM introduces disparity, scale and pose optimization (DSPO) to optimize camera poses, DROID disparities, and the scale and shift of monocular depth priors, while using the aligned prior to supplement regions lacking multi-view depth~\cite{sandstromSplatSLAMGloballyOptimized2024}. DROID-Splat \cite{homeyerDROIDSplatCombiningEndtoend2024} also supports monocular depth priors, but primarily treats them as an alternative source of depth supervision rather than using them to stabilize DROID depth. Although effective for RGB imagery, these strategies are less suitable for thermal data, where weak texture and non-uniform radiometric responses can produce spatially inconsistent monocular priors and insufficient multi-view constraints. Consequently, global scale--shift alignment and direct prior-based completion may introduce erroneous geometry into Gaussian mapping. To address these limitations, TOM-GS iteratively selects geometrically consistent prior estimates to stabilize the DROID depth, without directly filling regions lacking multi-view depth support.

\subsection{Monocular Depth Estimation}

Recent advances in learning-based monocular depth estimation have significantly improved depth prediction accuracy through large-scale training. Representative methods include ZoeDepth~\cite{bhatZoeDepthZeroshotTransfer2023}, Metric3D~\cite{yinMetric3DZeroshotMetric2023}, Depth Anything V2~\cite{yangDepthAnythingV22024}, and Lotus~\cite{heLotus2025}. These approaches provide dense geometric priors from a single image and have demonstrated strong generalization across diverse environments.

For thermal imagery, monocular depth estimation is particularly valuable due to the limited texture and weak geometric cues commonly observed in thermal scenes. While recent studies have explored thermal depth prediction~\cite{shinMaximizingSelfsupervisionThermal2022}, monocular depth priors have rarely been incorporated into Gaussian-based thermal reconstruction to provide geometric supervision and improve depth consistency.

\section{Thermal Odometry and Mapping with Depth-Guided GS}

\begin{figure*}[t]
    \centering
    \includegraphics[width=\textwidth]{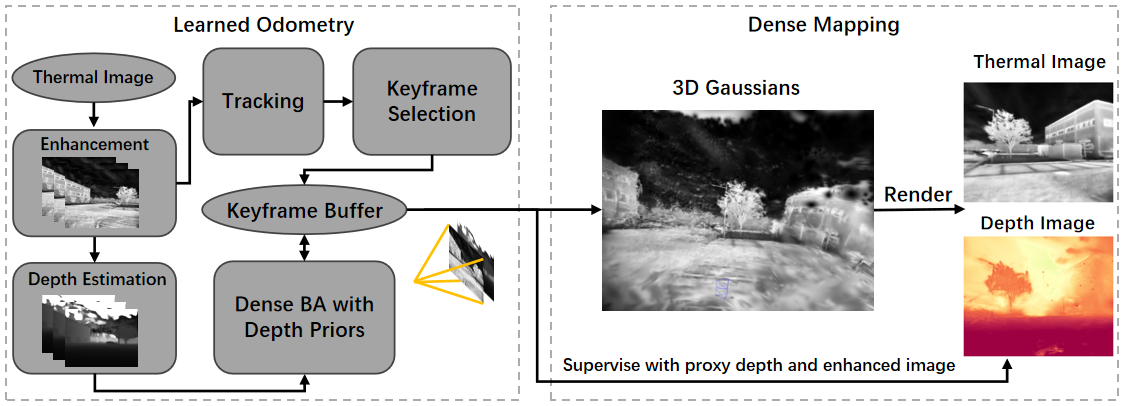}
\caption{Overview of the proposed GS-based SLAM system with a learning-based odometry module. Thermal images are first converted to 8-bit using an image enhancement module.
The resulting grayscale images are fed to a depth prediction network to obtain monocular depth maps and to the learning-based odometry module to estimate camera motion.
The odometry module tracks each frame against selected keyframes, performs keyframe selection, and refines keyframe poses and dense depth maps by fusing learned depth priors. The refined keyframes and depth estimates are finally passed to the 3D GS mapping module, which optimizes the 3D Gaussians representing the scene.
    }
    \label{fig:flowchart}
\end{figure*}

Given a sequence of monocular thermal infrared images, our objective is to estimate the camera motion (up to scale) and reconstruct the environment using Gaussian Splatting. Because many thermal cameras output 14- or 16-bit images, our pipeline includes a thermal image enhancement module. We do not consider Non-Uniformity Correction (NUC), as the missing or duplicated frames caused by NUC do not affect motion estimation when only image data is used. The proposed learning-based GS SLAM framework for monocular thermal imagery is illustrated in Fig.~\ref{fig:flowchart}.

To ensure robust odometry and decouple it from mapping, we decompose the proposed GS-based SLAM system into two modules: a learning-based odometry frontend and a GS-based mapping backend.
The odometry module estimates camera motion by combining dense image matches with learned depth priors. The GS mapping module uses the estimated poses, enhanced thermal images, and refined depth predictions to initialize Gaussian primitives and optimize their parameters through photometric and geometric losses.

\subsection{Learned Thermal Odometry with Depth Prior}





The odometry module takes a stream of enhanced thermal infrared images as input and estimates both the camera trajectory and a dense 3D structure for keyframes. The Thermal Infrared Odometry (TIO) system is composed of three main components: a recurrent convolutional module, ConvGRU~\cite{teedDROIDSLAMDeepVisual2021}, a dense bundle adjustment (DBA) layer~\cite{teedDROIDSLAMDeepVisual2021}, and a disparity and scale optimization (DSOpt) layer. The ConvGRU predicts optical flow updates conditioned on camera poses and dense depth estimates. These dense depths, referred to as DROID depth, are propagated and refined throughout the pipeline. The DBA layer jointly optimizes camera poses and dense depths using these flow corrections. The DSOpt layer optimizes the disparities and scales of keyframes, fusing DROID depth with monocular depth priors.

To represent the trajectory and map, TIO maintains a co-visibility graph $G(V,E)$ for keyframes. Keyframes are selected based on the magnitude of optical flow between frames. Each vertex $V_i$ corresponds to a keyframe $I_i$ and stores its pose $\mathbf{T}_i$ in a world frame and its dense inverse depth map $d_i$. An edge $(i, j)\in E$ indicates that frames $I_i$ and $I_j$ have an overlapping view. Pixel coordinates $\mathbf{p}_i \in \mathbb{R}^{H\times W \times 2}$ in $I_i$ are projected into $I_j$ to obtain $\mathbf{p}_{ij}$:
\begin{equation}
\mathbf p_{ij} = \Pi(\mathbf{T}_{ij} \Pi^{-1} (\mathbf{p}_i, d_i)), \ 
\mathbf{T}_{ij} = \mathbf{T}_j^{-1} \mathbf{T}_i,
\end{equation}
where $\Pi$ is the projection function that maps 3D points to image coordinates.
Its inverse $\Pi^{-1}$ back-projects pixel coordinates $\mathbf{p}_i$ with inverse depths $d_i$ into 3D.


For each frame, TIO extracts correlation and context features using two residual blocks. The correlation features capture visual similarity between pixels in images $I_i$ and $I_j$, while the context features help reject incorrect correspondences arising from dynamic regions or large moving objects. For each edge $(i,j)$, the convolutional GRU takes the correlation features and the flow residuals $\mathbf{p}_{ij} - \mathbf{p}_i$ to estimate a flow correction $\mathbf{r}_{ij}\in\mathbb{R}^{H\times W \times 2}$ and a confidence map $\mathbf{w}_{ij}\in\mathbb{R}^{H\times W \times 2}_+$. The DBA then refines both camera poses and dense depths over the co-visibility graph using the classic reprojection error objective
\begin{subequations} 
\begin{equation}
L(\mathbf{T}, d) = \sum_{i,j} \left\| \mathbf{p}_{ij} + \mathbf{r}_{ij} - \Pi(\mathbf{T}_{ij}\Pi^{-1}(\mathbf{p}_i, d_i)) \right\|^2_{\boldsymbol{\Sigma}_{ij}},
\end{equation}
\begin{equation}
\boldsymbol{\Sigma}_{ij} = \mathrm{diag}(\mathbf{w}_{ij}),
\end{equation}
\end{subequations}
where $\left\|\cdot\right\|_\Sigma$ denotes the Mahalanobis distance. To limit computation, older edges are gradually pruned from the graph. The first camera pose is fixed to eliminate gauge freedom.

To ensure geometric consistency, we first apply multi-view consistency filtering~\cite{sandstromSplatSLAMGloballyOptimized2024, zhangGlORIESLAMGloballyOptimized2024} to the DROID depth, yielding a valid depth mask $M_i^{d}$. Multi-view filtering removes inconsistent estimates but does not further constrain the retained depth. We therefore incorporate a monocular depth prior into the depth refinement stage to provide complementary geometric guidance, stabilizing the optimized depth and subsequent Gaussian mapping.


Unlike Splat-SLAM \cite{sandstromSplatSLAMGloballyOptimized2024}, which estimates both scale and shift, we estimate only a global scale to align the monocular prior with the DROID depth.
We ignore the shift for several reasons. First, many mono depth estimators, e.g., ZoeDepth \cite{bhatZoeDepthZeroshotTransfer2023} and Metric3D \cite{yinMetric3DZeroshotMetric2023}, provide metric depth that is related to DROID depth by a scale factor.
Second, depth errors caused by NUC are spatially varying, making a single global shift not useful.

To robustly estimate the scale, we adopt an iterative procedure.
The valid DROID depth mask $M_i^{d}$ is first used to initialize the prior mask, i.e., $M_i^{p(0)}=M_i^{d}$. The prior mask is then iteratively refined according to the alignment residual. At iteration $k$, the scale is estimated over the current prior mask $M_i^{p(k)}$ as
\begin{equation}
\theta_i^{(k)}
=
\frac{
\operatorname{median}_{(u,v)\in M_i^{p(k)}}
\left(D_i(u,v)\right)
}{
\operatorname{median}_{(u,v)\in M_i^{p(k)}}
\left(D_i^{\mathrm{mono}}(u,v)\right)
},
\end{equation}
where unfiltered DROID depth $D_i=d_i^{-1}$. The relative depth residual is computed
as
\begin{equation}
r_i^{(k)}(u,v)
=
\frac{
\left|
\theta_i^{(k)}D_i^{\mathrm{mono}}(u,v)-D_i(u,v)
\right|
}{
D_i(u,v)
},
\end{equation}
and the prior mask is updated by
\begin{equation}
M_i^{p(k+1)}
=
\left\{
(u,v)\in M_i^{p(k)}
\mid
r_i^{(k)}(u,v)<\tau
\right\}.
\end{equation}
Here, $\tau$ is the relative residual threshold and is set to $0.15$. The iteration terminates when the estimated scale converges, no valid pixels remain, or the maximum number of iterations is reached. The resulting scale $\theta_i$ and prior mask $M_i^p$ are then incorporated into the subsequent depth refinement objective:
\begin{align}
\mathcal{L}(d,\theta)
=
&
\sum_{(i,j)\in E}
\left\|
\mathbf{p}_{ij}
+\mathbf{r}_{ij}
-
\Pi
\!\left(
\mathbf{T}_{ij}
\Pi^{-1}
(\mathbf{p}_i,d_i)
\right)
\right\|_{\boldsymbol{\Sigma}_{ij}}^2
\nonumber\\
&
+\lambda_d
\sum_{i\in V}
\left\|
M_i^p\odot
\left(
d_i-\frac{1}{\theta_iD_i^{\mathrm{mono}}}
\right)
\right\|_2^2,
\end{align}
where $\odot$ denotes element-wise multiplication. Following Splat-SLAM, $\lambda_d$ is set to $0.01$ to balance the two terms.
Each optimization iteration consists of a depth scale refinement step followed by a DBA update.

\subsection{Depth-Guided Gaussian Splatting}

Given the estimated camera poses and the dense depth maps of keyframes, the mapping module constructs a scene representation using Gaussian Splatting. To improve efficiency and reduce redundancy, keyframes are further filtered using the co-visibility criteria from MonoGS~\cite{matsukiGaussian2024}. Although GS supports pose refinement during mapping, we disable this to avoid inconsistency with the odometry poses.

The scene is represented as a collection of 3D Gaussians. Each Gaussian $\mathcal{G}_i$  is characterized by its color $c_i$, opacity $o_i \in {\mathbb{R}}^1$, center position $\boldsymbol{\mu}_{i} \in {\mathbb{R}}^3$, and a scaling vector $\mathbf{s}_i\in {\mathbb{R}}^3$. For simplicity, we directly use grayscale intensity as the Gaussian color $c_i$. The rotation matrix $\mathbf{R}_i$ and the scaling matrix 
$\mathbf{S}_i = \text{diag}(s_{i1},s_{i2},s_{i3})$
together define the covariance of a Gaussian $\mathcal{G}_i$ as 
$\boldsymbol{\Sigma}_{W, i} = 
\mathbf R_i \mathbf S_i \mathbf S_i^T \mathbf R_i^T$. A 3D Gaussian influences a point $\mathbf{x} \in {\mathbb{R}}^3$ by
\begin{equation}
\mathcal{G}_i(\mathbf x) = o_i \exp[-0.5 (\mathbf x- \boldsymbol{\mu}_i)^T \boldsymbol{\Sigma}_{W,i}^{-1} (\mathbf x- \boldsymbol{\mu}_i)].
\end{equation} \
To render an image from a set of Gaussians, we first sort the Gaussians by their center depth. Each 3D Gaussian is then projected onto the image plane, forming a 2D Gaussian $\mathcal{G}_i^\prime$ with image-plane center $\boldsymbol{\mu}_i^\prime$ and covariance $\boldsymbol{\Sigma}_i^\prime$
\begin{equation}
\boldsymbol{\mu}_i^\prime = \boldsymbol{\pi} (\mathbf{T}_{CW} \cdot \boldsymbol{\mu}_i),
\end{equation}
\begin{equation}
\boldsymbol{\Sigma}_i^\prime = \mathbf{J} \mathbf{R}_{CW} \boldsymbol{\Sigma}_{W,i} \mathbf{R}_{CW}^T \mathbf{J}^T,
\end{equation}
where $\mathbf{J} = \frac{\partial \boldsymbol{\pi} (\mathbf{p}_C)}{\partial \mathbf{p}_C}$ is the Jacobian of the projection and $\mathbf{T}_{CW}$ transforms world coordinates into the camera frame. The projection $\boldsymbol{\pi}$ is defined by
\begin{equation}
\mathbf{W} = \mathbf{T}_{ndc}^{pix} \cdot \mathbf{T}_{cam}^{ndc},
\end{equation}
where $\mathbf{T}_{cam}^{ndc}$ maps camera coordinates to normalized device coordinates (NDC), and $\mathbf{T}_{ndc}^{pix}$ maps NDC to pixel coordinates.
A 2D Gaussian $\mathcal{G}_i^\prime$ contributes to a pixel $\mathbf{x}^\prime \in {\mathbb{R}}^2$ as
\begin{equation}
\mathcal{G}_i^\prime(\mathbf x^\prime) = o_i \exp[- 0.5 (\mathbf x^\prime - \boldsymbol{\mu}_i^\prime)^T \boldsymbol{\Sigma}_i^{\prime -1} (\mathbf {x}^\prime - \boldsymbol{\mu}_i^\prime)].
\end{equation} \
To compute the final pixel intensity at $\mathbf{p} = (u,v)$, all $n$ Gaussians sorted by depth are composited front-to-back using volumetric alpha blending:
\begin{equation}
c(\mathbf{p}) = \sum_{i=1}^n c_i \mathcal{G}_i^\prime(\mathbf{p})
	\prod_{j=1}^{i-1} (1- \mathcal{G}_j^\prime(\mathbf{p})).
\end{equation} \

In practice, the image is partitioned into 16$\times$16 tiles. For each tile, we identify all projected 2D Gaussians whose centers fall inside the tile and then perform alpha blending using only the Gaussians intersecting the tile. \

Depth rendering follows the same formulation. To get the depth of a pixel, $D(\mathbf{x})$, we simply replace $c_i$ with the 2D Gaussian’s depth $z_i = [\mathbf{R}_{CW} \boldsymbol{\mu}_i + \mathbf{p}_{CW}]_z$.

During mapping optimization, the grayscale images and proxy depths from the odometry module supervise the learnable GS parameters. The overall loss function comprises a color loss $\mathcal{L}_c$, a structural similarity loss $\mathcal{L}_{SSIM}$, and a depth loss $\mathcal{L}_d$,
\begin{equation}
\mathcal{L} = (1 -\alpha)\mathcal{L}_c + \alpha \mathcal{L}_{SSIM} + \beta \mathcal{L}_d,
\end{equation}
with empirical weights $\alpha=0.5$ and $\beta=0.2$. The terms $\mathcal{L}_c$ and $\mathcal{L}_{SSIM}$ measure the photometric and structural differences between rendered and observed images, while $\mathcal{L}_d$ is the mean $L_1$ error between the rendered depth and the proxy depth.



To improve the geometric supervision of Gaussian splatting mapping, we use the DSOpt-refined depth as the proxy depth. Unlike Splat-SLAM, the monocular prior depth is not used to provide supervision over all regions lacking valid DROID depths, as the mono depth predictions are not always reliable.
With the proxy depth, for each keyframe, Gaussians are spawned by sampling depths from the proxy frame. For Gaussian densification, the AbsGS strategy~\cite{yeAbsGSRecoveringFine2024} is adopted to insert new Gaussians based on the magnitude of the gradient with respect to the 2D Gaussian positions. While the absolute gradient guides densification, the signed gradient is retained for backpropagation. To prune redundant Gaussians, Gaussians with low opacity and large spatial extent are removed following the approach used in MonoGS and SplaTAM. Specifically, Gaussians with opacity below a threshold of 0.05 are removed. In addition, newly generated Gaussians that are not observed in at least three previous frames are also discarded.

\section{EXPERIMENTS}


\definecolor{colOurs}{RGB}{228,26,28}
\definecolor{colLDSO}{RGB}{55,126,184}
\definecolor{colDROID}{RGB}{77,175,74}
\definecolor{colDPVO}{RGB}{152,78,163}
\definecolor{colNeRFVO}{RGB}{255,127,0}
\definecolor{colTMD}{RGB}{166,86,40}
\definecolor{colPhoto}{RGB}{247,129,191}
\definecolor{colMonoGS}{RGB}{153,153,153}
\definecolor{colGlORIE}{RGB}{200,180,0}
\definecolor{colORB3}{RGB}{0,0,0}

%

\pgfplotsset{
  myaxis/.style={
    only marks,
    width=0.9\textwidth,
    height=0.25\textwidth,
    ymin=0.009, ymax=28,
    xmin=0.5, xmax=10.5,
    yminorticks=true,
    xtick={1,2,3,4,5,6,7,8,9,10},
    grid=major,
    ymajorgrids=true,
    xmajorgrids=false,
    grid style={dashed, gray!30},
    axis on top, 
    ylabel={},
    clip=false,
  },
}
\begin{figure*}[t]
\centering
\begin{tikzpicture}

\begin{semilogyaxis}[
  myaxis, name=topaxis,
  xticklabels={,,,,,,,,,},
  xlabel={},
  legend style={draw=none},
]

\fill[blue!5]   (axis cs:0.5,0.009) rectangle (axis cs:7.5,28);
\fill[orange!8] (axis cs:7.5,0.009) rectangle (axis cs:9.5,28);
\fill[gray!10]  (axis cs:9.5,0.009) rectangle (axis cs:10.5,28);

\draw[gray!40,densely dashed,line width=0.5pt] (axis cs:1.5,0.009)--(axis cs:1.5,28);
\draw[gray!40,densely dashed,line width=0.5pt] (axis cs:2.5,0.009)--(axis cs:2.5,28);
\draw[gray!40,densely dashed,line width=0.5pt] (axis cs:3.5,0.009)--(axis cs:3.5,28);
\draw[gray!40,densely dashed,line width=0.5pt] (axis cs:4.5,0.009)--(axis cs:4.5,28);
\draw[gray!40,densely dashed,line width=0.5pt] (axis cs:5.5,0.009)--(axis cs:5.5,28);
\draw[gray!40,densely dashed,line width=0.5pt] (axis cs:6.5,0.009)--(axis cs:6.5,28);
\draw[gray!40,densely dashed,line width=0.5pt] (axis cs:7.5,0.009)--(axis cs:7.5,28);
\draw[gray!40,densely dashed,line width=0.5pt] (axis cs:8.5,0.009)--(axis cs:8.5,28);
\draw[gray!40,densely dashed,line width=0.5pt] (axis cs:9.5,0.009)--(axis cs:9.5,28);

\node[font=\scriptsize\bfseries,text=blue!60!black,inner sep=1.5pt,rounded corners=1pt]
  at (axis cs:4,15) {Indoor};
\node[font=\scriptsize\bfseries,text=orange!70!black,inner sep=1.5pt,rounded corners=1pt]
  at (axis cs:8.5,15) {Outdoor};
\node[font=\scriptsize\bfseries,text=gray!60!black,inner sep=1.5pt,rounded corners=1pt]
  at (axis cs:10,15) {Mean};

\addplot[solid,line width=0.6pt,color=colORB3,
  mark=o, mark size=1.5pt,
  mark options={draw=colORB3,fill=none,line width=0.6pt}]
  coordinates {
    (0.595,0.02)(1.595,0.06)(5.595,2.64)
    (6.595,0.05)(7.595,0.33)(8.595,0.22)(9.595,0.56)};

\addplot[solid,line width=0.6pt,color=colDROID,
  mark=triangle, mark size=1.5pt,
  mark options={draw=colDROID,fill=none,line width=0.6pt}]
  coordinates {
    (0.685,0.02)(1.685,0.06)(2.685,0.02)(3.685,1.53)(4.685,1.41)
    (5.685,0.33)(6.685,0.03)(7.685,0.15)(8.685,0.13)(9.685,0.41)};

\addplot[solid,line width=0.6pt,color=colGlORIE,
  mark=square, mark size=2pt,
  mark options={draw=colGlORIE,fill=none,line width=0.6pt}]
  coordinates {
    (0.775,0.02)(1.775,0.06)(3.775,1.54)(4.775,1.46)
    (6.775,2.11)(7.775,0.19)(8.775,0.15)(9.775,0.62)};

\addplot[only marks,solid,line width=0.6pt,color=colLDSO,
  mark=+, mark size=2pt,
  mark options={draw=colLDSO,fill=none,line width=0.6pt}]
  coordinates {
    (0.865,0.03)(1.865,0.64)(2.865,1.40)
    (5.865,5.08)(6.865,0.15)(7.865,0.39)(8.865,0.28)(9.865,1.14)};

\addplot[only marks,solid,line width=0.6pt,color=colPhoto,
  mark=triangle, mark size=1.5pt,
  mark options={rotate=180,draw=colPhoto,fill=none,line width=0.6pt}]
  coordinates {
    (0.955,0.02)(1.955,0.06)(2.955,1.34)
    (5.955,5.62)(6.955,0.36)(7.955,0.18)(8.955,0.24)(9.955,1.12)};

\addplot[only marks,solid,line width=0.6pt,color=colDPVO,
  mark=diamond, mark size=2pt,
  mark options={draw=colDPVO,fill=none,line width=0.6pt,rotate=90}]
  coordinates {
    (1.045,0.11)(2.045,0.16)(3.045,0.16)(4.045,1.54)(5.045,1.40)
    (6.045,1.03)(7.045,0.37)(8.045,0.61)(9.045,0.19)(10.045,0.62)};

\addplot[only marks,solid,line width=0.6pt,color=colNeRFVO,
  mark=triangle, mark size=2pt,
  mark options={draw=colNeRFVO,line width=0.6pt,rotate=90}]
  coordinates {(1.135,0.13)(6.135,9.27)(9.135,0.14)(10.135,3.18)};

\addplot[only marks,solid,line width=0.6pt,color=colTMD,
  mark=triangle, mark size=2pt,
  mark options={draw=colTMD,line width=0.6pt,rotate=270}]
  coordinates {
    (1.225,1.53)(2.225,1.44)(3.225,1.56)(4.225,1.54)(5.225,1.40)
    (6.225,10.44)(7.225,5.46)(8.225,8.57)(9.225,10.93)(10.225,4.76)};

\addplot[only marks,solid,line width=0.6pt,color=colMonoGS,
  mark=x, mark size=2.5pt,
  mark options={draw=colMonoGS,fill=none,line width=0.6pt}]
  coordinates {
    (1.315,1.57)(2.315,1.48)(3.315,1.58)
    (6.315,3.65)(7.315,4.82)(10.315,2.62)};

\addplot[solid,line width=1pt,color=colOurs,
  mark=diamond*, mark size=1.5pt,
  mark options={draw=colOurs,fill=colOurs,line width=0.6pt}]
  coordinates {
    (1.405,0.02)(2.405,0.06)(3.405,0.02)(4.405,1.55)(5.405,1.47)
    (6.405,0.17)(7.405,2.11)(8.405,0.21)(9.405,0.15)(10.405,0.23)};

\end{semilogyaxis}

\begin{semilogyaxis}[
  myaxis, name=botaxis,
  xticklabels={mocap\_easy,mocap\_medium,mocap\_difficult,mocap\_dark,mocap\_dark\_fast,indoor\_floor,gym,outdoor\_street,outdoor\_campus,Mean},
  xticklabel style={
    rotate=15,
    anchor=center,
    font=\scriptsize,
    inner sep=1pt,
    yshift=-10pt,
  },
  xlabel={},
  at={(topaxis.south west)}, anchor=north west, yshift=-0pt,
  legend style={
    at={(1.01,1.32)},
    anchor=west,
    font=\tiny,
    draw=black,
    fill=white,
    legend columns=1,
    legend cell align=left,
    row sep=2pt,
  },
]

\fill[blue!5]   (axis cs:0.5,0.009) rectangle (axis cs:7.5,28);
\fill[orange!8] (axis cs:7.5,0.009) rectangle (axis cs:9.5,28);
\fill[gray!10]  (axis cs:9.5,0.009) rectangle (axis cs:10.5,28);

\draw[gray!40,densely dashed,line width=0.5pt] (axis cs:1.5,0.009)--(axis cs:1.5,28);
\draw[gray!40,densely dashed,line width=0.5pt] (axis cs:2.5,0.009)--(axis cs:2.5,28);
\draw[gray!40,densely dashed,line width=0.5pt] (axis cs:3.5,0.009)--(axis cs:3.5,28);
\draw[gray!40,densely dashed,line width=0.5pt] (axis cs:4.5,0.009)--(axis cs:4.5,28);
\draw[gray!40,densely dashed,line width=0.5pt] (axis cs:5.5,0.009)--(axis cs:5.5,28);
\draw[gray!40,densely dashed,line width=0.5pt] (axis cs:6.5,0.009)--(axis cs:6.5,28);
\draw[gray!40,densely dashed,line width=0.5pt] (axis cs:7.5,0.009)--(axis cs:7.5,28);
\draw[gray!40,densely dashed,line width=0.5pt] (axis cs:8.5,0.009)--(axis cs:8.5,28);
\draw[gray!40,densely dashed,line width=0.5pt] (axis cs:9.5,0.009)--(axis cs:9.5,28);

\addplot[solid,line width=0.6pt,color=colORB3,
  mark=o, mark size=1.5pt,
  mark options={draw=colORB3,fill=none,line width=0.6pt}]
  coordinates {
    (0.595,0.16)(1.595,0.34)(3.595,0.11)(4.595,0.17)
    (5.595,0.25)(6.595,2.46)(8.595,0.14)(9.595,0.51)};
\addlegendentry{ORB3}

\addplot[solid,line width=0.6pt,color=colDROID,
  mark=triangle, mark size=1.5pt,
  mark options={draw=colDROID,fill=none,line width=0.6pt}]
  coordinates {
    (0.685,0.02)(1.685,0.07)(2.685,1.39)(3.685,0.08)(4.685,0.04)
    (5.685,6.14)(6.685,0.29)(7.685,4.00)(8.685,0.16)(9.685,1.35)};
\addlegendentry{DROID}

\addplot[solid,line width=0.6pt,color=colGlORIE,
  mark=square, mark size=2pt,
  mark options={draw=colGlORIE,fill=none,line width=0.6pt}]
  coordinates {
    (0.775,0.03)(1.775,0.07)(4.775,0.06)
    (6.775,1.02)(7.775,0.35)(8.775,0.23)(9.775,0.19)};
\addlegendentry{GlORIE}

\addplot[only marks,solid,line width=0.6pt,color=colLDSO,
  mark=+, mark size=2pt,
  mark options={draw=colLDSO,fill=none,line width=0.6pt}]
  coordinates {(0.865,1.12)(3.865,1.41)(8.865,0.27)(9.865,0.93)};
\addlegendentry{LDSO}

\addplot[only marks,solid,line width=0.6pt,color=colPhoto,
  mark=triangle, mark size=1.5pt,
  mark options={rotate=180,draw=colPhoto,fill=none,line width=0.6pt}]
  coordinates {
    (0.955,1.04)(1.955,0.86)(3.955,1.18)(4.955,0.86)
    (5.955,10.51)(6.955,3.49)(7.955,0.15)(8.955,0.45)(9.955,2.32)};
\addlegendentry{Photo}

\addplot[only marks,solid,line width=0.6pt,color=colDPVO,
  mark=diamond, mark size=2pt,
  mark options={draw=colDPVO,fill=none,line width=0.6pt,rotate=90}]
  coordinates {
    (1.045,0.24)(2.045,0.52)(3.045,1.12)(4.045,1.29)(5.045,0.21)
    (6.045,9.72)(7.045,5.50)(8.045,0.90)(9.045,0.18)(10.045,2.19)};
\addlegendentry{DPVO}

\addplot[only marks,solid,line width=0.6pt,color=colNeRFVO,
  mark=triangle, mark size=2pt,
  mark options={draw=colNeRFVO,line width=0.6pt,rotate=90}]
  coordinates {
    (1.135,0.24)(2.135,0.81)(3.135,6.19)(4.135,0.47)(5.135,0.34)
    (6.135,12.60)(7.135,0.92)(9.135,0.24)(10.135,2.72)};
\addlegendentry{NeRFVO}

\addplot[only marks,solid,line width=0.6pt,color=colTMD,
  mark=triangle, mark size=2pt,
  mark options={draw=colTMD,line width=0.6pt,rotate=270}]
  coordinates {
    (1.225,1.54)(2.225,1.45)(3.225,1.57)(4.225,1.54)(5.225,1.41)
    (6.225,10.45)(7.225,5.46)(8.225,19.23)(9.225,13.65)(10.225,6.26)};
\addlegendentry{TMD}

\addplot[only marks,solid,line width=0.6pt,color=colMonoGS,
  mark=x, mark size=2.5pt,
  mark options={draw=colMonoGS,fill=none,line width=0.6pt}]
  coordinates {(1.315,1.64)(3.315,1.58)(6.315,5.12)(9.315,2.78)};
\addlegendentry{MonoGS}

\addplot[solid,line width=1pt,color=colOurs,
  mark=diamond*, mark size=1.5pt,
  mark options={draw=colOurs,fill=colOurs,line width=0.6pt}]
  coordinates {
    (1.405,0.02)(2.405,0.07)(3.405,0.07)(4.405,0.08)(5.405,0.06)
    (6.405,9.66)(7.405,1.02)(8.405,0.43)(9.405,0.22)(10.405,0.35)};
\addlegendentry{Ours}

\end{semilogyaxis}

\node[rotate=90,anchor=center,font=\small]
  at ($(topaxis.west)!0.5!(botaxis.west)+(-1.2cm,0)$)
  {ATE RMSE (m) [log scale]};

\end{tikzpicture}

\vspace{-0.5em}
\caption{
Log-scale ATE RMSE on the RRXIO dataset for grayscale (top) and thermal (bottom) inputs. Tracking failures are omitted and excluded from the mean computation. For clarity, ATE values within each sequence are slightly offset horizontally. ORB: ORB-SLAM3; DROID: DROID-SLAM; GlORIE: GlORIE-SLAM; Photo: PhotoSLAM; TMD: ThermalMonoDepth.
}
\label{fig:rrxio_odom}

\end{figure*}
\definecolor{colOurs}{RGB}{228,26,28}
\definecolor{colLDSO}{RGB}{55,126,184}
\definecolor{colDROID}{RGB}{77,175,74}
\definecolor{colDPVO}{RGB}{152,78,163}
\definecolor{colNeRFVO}{RGB}{255,127,0}
\definecolor{colTMD}{RGB}{166,86,40}
\definecolor{colPhoto}{RGB}{247,129,191}
\definecolor{colMonoGS}{RGB}{153,153,153}
\definecolor{colGlORIE}{RGB}{200,180,0}
\definecolor{colORB3}{RGB}{0,0,0}

%

\pgfplotsset{
  myaxis/.style={
    width=0.9\textwidth,
    height=0.25\textwidth,
    only marks,
    ymin=0.025, ymax=22,
    xmin=0.5, xmax=15.5,
    yminorticks=true,
    xtick={1,2,3,4,5,6,7,8,9,10,11,12,13,14,15},
    grid=major,
    ymajorgrids=true,
    xmajorgrids=false,
    grid style={dashed, gray!30},
    axis on top, 
    ylabel={},
    clip=false,
  },
}

\begin{figure*}[t]
\centering
\begin{tikzpicture}

\begin{semilogyaxis}[
  myaxis, name=topaxis,
  xticklabels={,,,,,,,,,,,,,,},
  xlabel={},
  legend style={draw=none},
]

\fill[blue!5]   (axis cs:0.5,0.025)  rectangle (axis cs:10.5,22);
\fill[orange!8] (axis cs:10.5,0.025) rectangle (axis cs:14.5,22);
\fill[gray!8]   (axis cs:14.5,0.025) rectangle (axis cs:15.5,22);

\draw[gray!40,densely dashed,line width=0.5pt] (axis cs:1.5,0.025)--(axis cs:1.5,22);
\draw[gray!40,densely dashed,line width=0.5pt] (axis cs:2.5,0.025)--(axis cs:2.5,22);
\draw[gray!40,densely dashed,line width=0.5pt] (axis cs:3.5,0.025)--(axis cs:3.5,22);
\draw[gray!40,densely dashed,line width=0.5pt] (axis cs:4.5,0.025)--(axis cs:4.5,22);
\draw[gray!40,densely dashed,line width=0.5pt] (axis cs:5.5,0.025)--(axis cs:5.5,22);
\draw[gray!40,densely dashed,line width=0.5pt] (axis cs:6.5,0.025)--(axis cs:6.5,22);
\draw[gray!40,densely dashed,line width=0.5pt] (axis cs:7.5,0.025)--(axis cs:7.5,22);
\draw[gray!40,densely dashed,line width=0.5pt] (axis cs:8.5,0.025)--(axis cs:8.5,22);
\draw[gray!40,densely dashed,line width=0.5pt] (axis cs:9.5,0.025)--(axis cs:9.5,22);
\draw[gray!40,densely dashed,line width=0.5pt] (axis cs:10.5,0.025)--(axis cs:10.5,22);
\draw[gray!40,densely dashed,line width=0.5pt] (axis cs:11.5,0.025)--(axis cs:11.5,22);
\draw[gray!40,densely dashed,line width=0.5pt] (axis cs:12.5,0.025)--(axis cs:12.5,22);
\draw[gray!40,densely dashed,line width=0.5pt] (axis cs:13.5,0.025)--(axis cs:13.5,22);
\draw[gray!40,densely dashed,line width=0.5pt] (axis cs:14.5,0.025)--(axis cs:14.5,22);

\node[font=\scriptsize\bfseries,text=blue!60!black,inner sep=1.5pt,rounded corners=1pt]
  at (axis cs:5.5,12) {Indoor};
\node[font=\scriptsize\bfseries,text=orange!70!black,inner sep=1.5pt,rounded corners=1pt]
  at (axis cs:12.5,12) {Outdoor};
\node[font=\scriptsize\bfseries,text=gray!60!black,inner sep=1.5pt,rounded corners=1pt]
  at (axis cs:15,12) {Mean};

\addplot[solid,line width=0.6pt,color=colORB3,
  mark=o,mark size=1.5pt,mark options={draw=colORB3,fill=none,line width=0.6pt}]
  coordinates {(0.73,0.08)(1.73,0.04)(2.73,0.08)(6.73,0.32)(8.73,0.07)(9.73,0.18)(13.73,0.85)(14.73,0.23)};

\addplot[solid,line width=0.6pt,color=colDROID,
  mark=triangle,mark size=1.5pt,mark options={draw=colDROID,fill=none,line width=0.6pt}]
  coordinates {(0.79,0.09)(1.79,0.06)(2.79,0.46)(3.79,0.52)(4.79,0.64)(5.79,0.64)(6.79,0.06)(7.79,0.04)(8.79,0.11)(9.79,0.55)(10.79,0.36)(11.79,0.25)(12.79,2.38)(13.79,6.66)(14.79,0.92)};

\addplot[solid,line width=0.6pt,color=colGlORIE,
  mark=square,mark size=2pt,mark options={draw=colGlORIE,fill=none,line width=0.6pt}]
  coordinates {(0.85,0.09)(1.85,0.06)(2.85,0.11)(3.85,0.43)(4.85,0.71)(5.85,0.69)(6.85,0.06)(7.85,0.04)(8.85,0.11)(9.85,0.07)(10.85,0.37)(11.85,0.25)(12.85,2.02)(13.85,0.28)(14.85,0.38)};

\addplot[only marks,solid,line width=0.6pt,color=colLDSO,
  mark=+,mark size=2pt,mark options={draw=colLDSO,fill=none,line width=0.6pt}]
  coordinates {(0.91,0.52)(6.91,0.23)(7.91,0.65)(12.91,10.91)(14.91,3.08)};

\addplot[only marks,solid,line width=0.6pt,color=colPhoto,
  mark=triangle,mark size=1.5pt,mark options={rotate=180,draw=colPhoto,fill=none,line width=0.6pt}]
  coordinates {(0.97,0.46)(1.97,0.51)(6.97,0.06)(7.97,0.24)(8.97,0.25)(9.97,0.51)(10.97,0.36)(11.97,0.76)(12.97,11.51)(13.97,10.44)(14.97,2.51)};

\addplot[only marks,solid,line width=0.6pt,color=colDPVO,
  mark=diamond,mark size=2pt,mark options={draw=colDPVO,fill=none,line width=0.6pt,rotate=90}]
  coordinates {(1.03,0.09)(2.03,0.25)(3.03,0.17)(4.03,0.49)(5.03,0.62)(6.03,0.69)(7.03,0.06)(8.03,0.09)(9.03,0.24)(10.03,0.12)(11.03,0.54)(12.03,0.29)(13.03,4.32)(14.03,1.08)(15.03,0.65)};

\addplot[only marks,solid,line width=0.6pt,color=colNeRFVO,
  mark=triangle,mark size=2pt,mark options={draw=colNeRFVO,line width=0.6pt,rotate=90}]
  coordinates {(1.09,0.08)(2.09,0.12)(3.09,0.16)(4.09,0.78)(5.09,0.94)(6.09,1.66)(7.09,0.06)(8.09,0.04)(9.09,0.16)(10.09,4.44)(11.09,0.74)(12.09,0.37)(13.09,0.73)(14.09,0.38)(15.09,0.76)};

\addplot[only marks,solid,line width=0.6pt,color=colTMD,
  mark=triangle,mark size=2pt,mark options={draw=colTMD,line width=0.6pt,rotate=270}]
  coordinates {(1.15,0.66)(2.15,0.80)(3.15,0.78)(4.15,0.51)(5.15,0.75)(6.15,0.70)(7.15,0.42)(8.15,0.67)(9.15,0.61)(10.15,0.55)(11.15,1.73)(12.15,2.97)(13.15,3.05)(14.15,2.56)(15.15,1.20)};

\addplot[only marks,solid,line width=0.6pt,color=colMonoGS,
  mark=x,mark size=2.5pt,mark options={draw=colMonoGS,fill=none,line width=0.6pt}]
  coordinates {(1.21,0.58)(2.21,0.76)(3.21,0.69)(7.21,0.41)(8.21,0.63)(9.21,0.53)(10.21,0.49)(11.21,10.31)(12.21,7.93)(15.21,2.48)};

\addplot[solid,line width=1pt,color=colOurs,
  mark=diamond*, mark size=1.5pt,
  mark options={draw=colOurs,fill=colOurs,line width=0.6pt}]
  coordinates {(1.27,0.09)(2.27,0.06)(4.27,0.48)(5.27,0.72)(6.27,0.66)(7.27,0.06)(8.27,0.04)(10.27,0.06)(11.27,0.37)(12.27,0.26)(13.27,1.97)(14.27,0.40)(15.27,0.37)};

\end{semilogyaxis}

\begin{semilogyaxis}[
  myaxis, name=botaxis,
  xticklabels={
    in\_rob\_global, in\_unst\_global, in\_agg\_global,
    in\_rob\_dark,   in\_unst\_dark,   in\_agg\_dark,
    in\_rob\_local,  in\_unst\_local,  in\_agg\_local,
    in\_rob\_varying,
    out\_rob\_day1, out\_rob\_day2, out\_rob\_night1, out\_rob\_night2,
    Mean
  },
  xticklabel style={
    rotate=20,
    anchor=center,
    font=\scriptsize,
    inner sep=1pt,
    yshift=-15pt,
  },
  at={(topaxis.south west)}, anchor=north west, yshift=-0pt,
  legend style={
    at={(1.01,1.32)},
    anchor=west,
    font=\tiny,
    draw=black,
    fill=white,
    legend columns=1,
    legend cell align=left,
    row sep=2pt,
  },
]

\fill[blue!5]   (axis cs:0.5,0.025)  rectangle (axis cs:10.5,22);
\fill[orange!8] (axis cs:10.5,0.025) rectangle (axis cs:14.5,22);
\fill[gray!8]   (axis cs:14.5,0.025) rectangle (axis cs:15.5,22);

\draw[gray!40,densely dashed,line width=0.5pt] (axis cs:1.5,0.025)--(axis cs:1.5,22);
\draw[gray!40,densely dashed,line width=0.5pt] (axis cs:2.5,0.025)--(axis cs:2.5,22);
\draw[gray!40,densely dashed,line width=0.5pt] (axis cs:3.5,0.025)--(axis cs:3.5,22);
\draw[gray!40,densely dashed,line width=0.5pt] (axis cs:4.5,0.025)--(axis cs:4.5,22);
\draw[gray!40,densely dashed,line width=0.5pt] (axis cs:5.5,0.025)--(axis cs:5.5,22);
\draw[gray!40,densely dashed,line width=0.5pt] (axis cs:6.5,0.025)--(axis cs:6.5,22);
\draw[gray!40,densely dashed,line width=0.5pt] (axis cs:7.5,0.025)--(axis cs:7.5,22);
\draw[gray!40,densely dashed,line width=0.5pt] (axis cs:8.5,0.025)--(axis cs:8.5,22);
\draw[gray!40,densely dashed,line width=0.5pt] (axis cs:9.5,0.025)--(axis cs:9.5,22);
\draw[gray!40,densely dashed,line width=0.5pt] (axis cs:10.5,0.025)--(axis cs:10.5,22);
\draw[gray!40,densely dashed,line width=0.5pt] (axis cs:11.5,0.025)--(axis cs:11.5,22);
\draw[gray!40,densely dashed,line width=0.5pt] (axis cs:12.5,0.025)--(axis cs:12.5,22);
\draw[gray!40,densely dashed,line width=0.5pt] (axis cs:13.5,0.025)--(axis cs:13.5,22);
\draw[gray!40,densely dashed,line width=0.5pt] (axis cs:14.5,0.025)--(axis cs:14.5,22);

\addplot[solid,line width=0.6pt,color=colORB3,
  mark=o,mark size=1.5pt,mark options={draw=colORB3,fill=none,line width=0.6pt}]
  coordinates {(3.73,0.04)(9.73,0.04)(13.73,1.20)(14.73,0.43)};
\addlegendentry{ORB3}

\addplot[solid,line width=0.6pt,color=colDROID,
  mark=triangle,mark size=1.5pt,mark options={draw=colDROID,fill=none,line width=0.6pt}]
  coordinates {(0.79,0.67)(1.79,0.08)(2.79,0.08)(3.79,0.06)(4.79,0.06)(5.79,0.09)(6.79,0.05)(7.79,0.04)(8.79,0.07)(9.79,0.03)(10.79,5.91)(11.79,1.94)(12.79,3.20)(13.79,0.24)(14.79,0.89)};
\addlegendentry{DROID}

\addplot[solid,line width=0.6pt,color=colGlORIE,
  mark=square,mark size=2pt,mark options={draw=colGlORIE,fill=none,line width=0.6pt}]
  coordinates {(0.85,0.08)(1.85,0.31)(2.85,0.08)(3.85,0.07)(4.85,0.06)(5.85,0.09)(6.85,0.06)(7.85,0.04)(8.85,0.56)(9.85,0.03)(10.85,1.25)(11.85,0.24)(12.85,1.97)(13.85,0.28)(14.85,0.37)};
\addlegendentry{GlORIE}

\addplot[only marks,solid,line width=0.6pt,color=colLDSO,
  mark=+,mark size=2pt,mark options={draw=colLDSO,fill=none,line width=0.6pt}]
  coordinates {(3.91,0.45)(6.91,0.51)(7.91,0.69)(14.91,0.55)};
\addlegendentry{LDSO}

\addplot[only marks,solid,line width=0.6pt,color=colPhoto,
  mark=triangle,mark size=1.5pt,mark options={rotate=180,draw=colPhoto,fill=none,line width=0.6pt}]
  coordinates {(3.97,0.29)(9.97,0.20)(10.97,0.41)(11.97,1.46)(12.97,0.96)(13.97,1.31)(14.97,0.77)};
\addlegendentry{Photo}

\addplot[only marks,solid,line width=0.6pt,color=colDPVO,
  mark=diamond,mark size=2pt,mark options={draw=colDPVO,fill=none,line width=0.6pt,rotate=90}]
  coordinates {(1.03,0.51)(2.03,0.67)(3.03,0.76)(4.03,0.14)(5.03,0.19)(6.03,0.61)(7.03,0.07)(8.03,0.09)(9.03,0.16)(10.03,0.07)(11.03,0.83)(12.03,1.36)(13.03,1.24)(14.03,0.95)(15.03,0.55)};
\addlegendentry{DPVO}

\addplot[only marks,solid,line width=0.6pt,color=colNeRFVO,
  mark=triangle,mark size=2pt,mark options={draw=colNeRFVO,line width=0.6pt,rotate=90}]
  coordinates {(1.09,0.18)(2.09,2.71)(3.09,1.81)(4.09,0.05)(5.09,0.09)(6.09,0.10)(7.09,0.05)(8.09,0.05)(9.09,0.09)(10.09,0.04)(11.09,1.75)(12.09,1.81)(13.09,1.16)(14.09,1.33)(15.09,0.80)};
\addlegendentry{NeRFVO}

\addplot[only marks,solid,line width=0.6pt,color=colTMD,
  mark=triangle,mark size=2pt,mark options={draw=colTMD,line width=0.6pt,rotate=270}]
  coordinates {(1.15,0.27)(2.15,0.54)(3.15,0.65)(4.15,0.42)(5.15,0.38)(6.15,0.70)(7.15,0.29)(8.15,0.38)(9.15,0.60)(10.15,0.46)(11.15,1.74)(12.15,2.98)(13.15,3.05)(14.15,2.56)(15.15,1.07)};
\addlegendentry{TMD}

\addplot[only marks,solid,line width=0.6pt,color=colMonoGS,
  mark=x,mark size=2.5pt,mark options={draw=colMonoGS,fill=none,line width=0.6pt}]
  coordinates {(1.21,0.69)(3.21,0.75)(4.21,0.48)(5.21,0.72)(7.21,0.35)(8.21,0.52)(10.21,0.31)(11.21,6.13)(12.21,7.55)(13.21,5.29)(14.21,6.61)(15.21,2.67)};
\addlegendentry{MonoGS}

\addplot[solid,line width=1pt,color=colOurs,
  mark=diamond*, mark size=1.5pt,
  mark options={draw=colOurs,fill=colOurs,line width=0.6pt}]
  coordinates {(1.27,0.07)(2.27,0.33)(3.27,0.09)(4.27,0.05)(5.27,0.06)(6.27,0.63)(7.27,0.06)(8.27,0.04)(9.27,0.51)(10.27,0.03)(11.27,1.24)(12.27,0.24)(13.27,2.06)(14.27,0.29)(15.27,0.41)};
\addlegendentry{Ours}

\end{semilogyaxis}

\node[rotate=90,anchor=center,font=\small]
  at ($(topaxis.west)!0.5!(botaxis.west)+(-1.2cm,0)$)
  {ATE RMSE (m) [log scale]};

\end{tikzpicture}

\vspace{-0.5em}
\caption{
Log-scale ATE RMSE on the VIVID dataset for RGB (top) and thermal (bottom) inputs. Tracking failures are omitted and excluded from the mean computation. For clarity, ATE values within each sequence are slightly offset horizontally. ORB: ORB-SLAM3; DROID: DROID-SLAM; GlORIE: GlORIE-SLAM; Photo: PhotoSLAM; TMD: ThermalMonoDepth.
}
\label{fig:vivid_odom}

\end{figure*}

We evaluate the proposed monocular thermal odometry and Gaussian Splatting–based mapping framework (TOM-GS) on two public benchmarks, RRXIO~\cite{doerRadarInertialOdometry2020} and VIVID~\cite{leeViViDVisionVisibility2022}, focusing on both tracking accuracy and reconstruction quality. 

The RRXIO dataset contains nine sequences captured in indoor and outdoor scenes using a monochrome camera (1280$\times$1024) and a thermal camera (640$\times$512). Indoor sequences include accurate ground truth from a motion capture system, while outdoor reference trajectories are generated using visual–inertial SLAM. During preprocessing, grayscale images from both cameras are converted to three channels and undistorted. The VIVID dataset contains 14 sequences covering indoor and outdoor environments. It includes RGB-D data (640$\times$480) and thermal images captured with a FLIR A65 camera (640$\times$512). Since VIVID thermal data are provided in 14-bit format, the image enhancement module in Fig.~\ref{fig:flowchart} is required before feeding the images to odometry and mapping networks.

\begin{table*}[t]
\centering
\caption{ Rendering results on the RRXIO sequences using NERF-VO, PhotoSLAM, MonoGS, GLORIE-SLAM, and our TOM-GS. Photo and GLORIE denote PhotoSLAM and GLORIE-SLAM, respectively. Each cell shows Gray\ \textbar\  Thermal values. '-' indicates failure. Average values are computed over successful runs. Best is \textbf{bolded}; second-best is \underline{underlined}.}
\label{tab:rrxio_render}
\resizebox{\textwidth}{!}{
\begin{tabular}{c|c|*{9}{c}|c}
\Xhline{1.5pt}
Method & Metric & easy & medium & diff. & dark & fast & floor & gym & street & campus & Avg. \\
\Xhline{1.5pt}
\multirow{2}{*}{NeRFVO}
 & \setcounter{showsep}{1} PSNR$\uparrow$  & \dualval{28.27}{28.55} & \dualval{-}{28.28} & \dualval{-}{28.25} & \dualval{-}{28.19} & \dualval{-}{28.29} & \dualval{27.97}{28.02} & \dualval{-}{28.36} & \dualval{-}{-} & \dualval{29.23}{29.36} & \dualval{\textbf{28.49}}{28.41} \\
 & LPIPS$\downarrow$ & \dualval{0.60}{0.61} & \dualval{-}{0.65} & \dualval{-}{0.56} & \dualval{-}{0.65} & \dualval{-}{0.62} & \dualval{0.80}{0.64} & \dualval{-}{0.65} & \dualval{-}{-} & \dualval{0.47}{0.44} & \dualval{0.62}{0.60} \\
\hline
\multirow{1}{*}{Photo}
 & \setcounter{showsep}{1} PSNR$\uparrow$  & \dualval{26.83}{27.03} & \dualval{26.62}{29.97} & \dualval{16.73}{-} & \dualval{-}{24.59} & \dualval{-}{32.83} & \dualval{18.90}{24.79} & \dualval{24.34}{29.93} & \dualval{19.90}{24.02} & \dualval{25.98}{27.93} & \dualval{22.76}{\underline{27.64}} \\
\hline
\multirow{2}{*}{MonoGS}
 & \setcounter{showsep}{1} PSNR$\uparrow$  & \dualval{17.16}{23.50} & \dualval{17.65}{-} & \dualval{14.25}{-} & \dualval{-}{15.40} & \dualval{-}{-} & \dualval{7.21}{-} & \dualval{20.75}{25.84} & \dualval{-}{-} & \dualval{-}{-} & \dualval{15.40}{21.58} \\
 & LPIPS$\downarrow$ & \dualval{0.60}{0.62} & \dualval{0.57}{-} & \dualval{0.64}{-} & \dualval{-}{0.82} & \dualval{-}{-} & \dualval{0.81}{-} & \dualval{0.45}{0.58} & \dualval{-}{-} & \dualval{-}{-} & \dualval{0.61}{0.67} \\
\hline
\multirow{2}{*}{GLORIE}
 & \setcounter{showsep}{1} PSNR$\uparrow$  & \dualval{17.87}{23.77} & \dualval{17.83}{21.67} & \dualval{-}{-} & \dualval{31.21}{-} & \dualval{31.18}{22.61} & \dualval{-}{-} & \dualval{19.86}{21.16} & \dualval{13.80}{18.91} & \dualval{18.36}{22.59} & \dualval{21.45}{21.78} \\
 & LPIPS$\downarrow$ & \dualval{0.62}{0.50} & \dualval{0.64}{0.57} & \dualval{-}{-} & \dualval{0.08}{-} & \dualval{0.07}{0.54} & \dualval{-}{-} & \dualval{0.60}{0.65} & \dualval{0.77}{0.61} & \dualval{0.59}{0.54} & \dualval{\underline{0.48}}{\underline{0.57}} \\
\hline
\multirow{2}{*}{Ours}
 & \setcounter{showsep}{1} PSNR$\uparrow$  & \dualval{28.55}{30.76} & \dualval{27.08}{27.90} & \dualval{26.34}{25.63} & \dualval{-}{28.33} & \dualval{-}{28.46} & \dualval{-}{-} & \dualval{28.23}{33.00} & \dualval{-}{-} & \dualval{-}{-} & \dualval{\underline{27.55}}{\textbf{29.01}} \\
 & LPIPS$\downarrow$ & \dualval{0.07}{0.22} & \dualval{0.10}{0.29} & \dualval{0.10}{0.42} & \dualval{-}{0.27} & \dualval{-}{0.30} & \dualval{-}{-} & \dualval{0.10}{0.25} & \dualval{-}{-} & \dualval{-}{-} & \dualval{\textbf{0.09}}{\textbf{0.29}} \\
\Xhline{1.5pt}
\end{tabular}
}
\end{table*}
\begin{table*}[t]
\centering
\caption{Rendering results on VIVID sequences using NERF-VO, PhotoSLAM, MonoGS, GLORIE-SLAM, and our TOM-GS. The dataset includes indoor (I) and outdoor (O) sequences, with slow robust (R), unstable (U), and aggressive (A) motion under various illumination: global light (G), lights off (Dark or Night), local flashlight (L), and flashlight varying from dark to local (V). Each cell shows RGB \textbar\  Thermal values. '-' indicates failure. Average values are computed over successful runs. Best is \textbf{bolded}; second-best is \underline{underlined}.}

\label{tab:vivid_render}
\resizebox{\textwidth}{!}{
\setlength{\tabcolsep}{1.5pt}
\begin{tabular}{c|c|*{14}{c}|c}
\Xhline{1.5pt}
Method & Metric & IRG & IUG & IAG & IRD & IUD & IAD & IRL & IUL & IAL & IRV & ORD1 & ORD2 & ORN1 & ORN2 & Avg. \\
\Xhline{1.5pt}
\multirow{2}{*}{NeRFVO}
 & \setcounter{showsep}{1} PSNR$\uparrow$
 & \dualval{28.03}{28.03} & \dualval{27.93}{28.00} & \dualval{28.04}{27.72} & \dualval{-}{27.59} & \dualval{-}{28.63} & \dualval{-}{27.61} & \dualval{28.18}{27.93} & \dualval{28.31}{27.67} & \dualval{28.17}{27.91} & \dualval{28.31}{27.68} & \dualval{28.13}{28.00} & \dualval{28.02}{28.00} & \dualval{-}{28.09} & \dualval{-}{28.10} & \dualval{\underline{28.12}}{\textbf{27.93}} \\
 & LPIPS$\downarrow$
 & \dualval{0.62}{0.36} & \dualval{0.65}{0.39} & \dualval{0.65}{0.35} & \dualval{-}{0.43} & \dualval{-}{0.28} & \dualval{-}{0.39} & \dualval{0.55}{0.25} & \dualval{0.55}{0.27} & \dualval{0.56}{0.30} & \dualval{0.52}{0.30} & \dualval{0.56}{0.67} & \dualval{0.56}{0.66} & \dualval{-}{0.63} & \dualval{-}{0.58} & \dualval{0.58}{\underline{0.42}} \\
\hline
\multirow{1}{*}{Photo}
 & \setcounter{showsep}{1} PSNR$\uparrow$
 & \dualval{17.23}{-} & \dualval{14.43}{-} & \dualval{18.96}{-} & \dualval{-}{24.74} & \dualval{-}{-} & \dualval{-}{-} & \dualval{20.07}{25.93} & \dualval{19.61}{-} & \dualval{24.51}{-} & \dualval{15.90}{25.11} & \dualval{20.50}{20.09} & \dualval{20.76}{19.66} & \dualval{22.39}{25.13} & \dualval{21.61}{24.43} & \dualval{19.64}{23.58} \\
\hline
\multirow{2}{*}{MonoGS}
 & \setcounter{showsep}{1} PSNR$\uparrow$
 & \dualval{17.18}{24.13} & \dualval{17.25}{-} & \dualval{15.51}{20.53} & \dualval{-}{24.00} & \dualval{-}{22.08} & \dualval{-}{-} & \dualval{19.12}{26.22} & \dualval{17.58}{24.05} & \dualval{16.38}{-} & \dualval{13.11}{23.89} & \dualval{21.48}{22.60} & \dualval{22.23}{23.47} & \dualval{-}{22.09} & \dualval{-}{22.31} & \dualval{17.76}{23.22} \\
 & LPIPS$\downarrow$
 & \dualval{0.47}{0.60} & \dualval{0.53}{-} & \dualval{0.59}{0.74} & \dualval{-}{0.55} & \dualval{-}{0.63} & \dualval{-}{-} & \dualval{0.47}{0.49} & \dualval{0.56}{0.54} & \dualval{0.65}{-} & \dualval{0.63}{0.52} & \dualval{0.43}{0.48} & \dualval{0.35}{0.43} & \dualval{-}{0.48} & \dualval{-}{0.50} & \dualval{0.52}{0.54} \\
\hline
\multirow{2}{*}{GLORIE}
 & \setcounter{showsep}{1} PSNR$\uparrow$
 & \dualval{16.04}{21.39} & \dualval{15.66}{20.12} & \dualval{15.19}{20.45} & \dualval{40.32}{22.06} & \dualval{40.17}{20.75} & \dualval{36.37}{21.59} & \dualval{17.84}{22.64} & \dualval{18.06}{23.13} & \dualval{17.07}{22.13} & \dualval{13.77}{24.06} & \dualval{15.68}{-} & \dualval{17.27}{-} & \dualval{18.79}{17.18} & \dualval{19.06}{17.44} & \dualval{21.52}{21.08} \\
 & LPIPS$\downarrow$
 & \dualval{0.53}{0.51} & \dualval{0.66}{0.55} & \dualval{0.70}{0.55} & \dualval{0.09}{0.46} & \dualval{0.09}{0.50} & \dualval{0.10}{0.50} & \dualval{0.52}{0.43} & \dualval{0.56}{0.43} & \dualval{0.65}{0.50} & \dualval{0.75}{0.42} & \dualval{0.63}{-} & \dualval{0.63}{-} & \dualval{0.39}{0.61} & \dualval{0.39}{0.60} & \dualval{\underline{0.48}}{0.50} \\
\hline
\multirow{2}{*}{Ours}
& \setcounter{showsep}{1} PSNR$\uparrow$
& \dualval{23.87}{27.16}
& \dualval{22.69}{25.72}
& \dualval{22.60}{26.97}
& \dualval{44.70}{27.95}
& \dualval{40.41}{28.45}
& \dualval{35.17}{25.79}
& \dualval{24.31}{29.91}
& \dualval{22.85}{30.27}
& \dualval{22.97}{28.44}
& \dualval{25.47}{29.21}
& \dualval{27.64}{14.35}
& \dualval{26.29}{21.87}
& \dualval{32.67}{24.94}
& \dualval{32.66}{25.97}
& \dualval{\textbf{28.88}}{\underline{26.22}} \\
& LPIPS$\downarrow$
& \dualval{0.17}{0.29}
& \dualval{0.24}{0.36}
& \dualval{0.27}{0.30}
& \dualval{0.03}{0.27}
& \dualval{0.06}{0.25}
& \dualval{0.08}{0.39}
& \dualval{0.20}{0.23}
& \dualval{0.25}{0.21}
& \dualval{0.32}{0.37}
& \dualval{0.25}{0.22}
& \dualval{0.10}{0.56}
& \dualval{0.11}{0.29}
& \dualval{0.07}{0.26}
& \dualval{0.06}{0.24}
& \dualval{\textbf{0.16}}{\textbf{0.30}} \\
 
\Xhline{1.5pt}
\end{tabular}
}
\end{table*}

\subsection{Odometry Evaluation}

For motion tracking, we compare ORB-SLAM3 \cite{camposORBSLAM3AccurateOpensource2021}, LDSO \cite{gaoLDSODirectSparse2018}, 
DROID-SLAM \cite{teedDROIDSLAMDeepVisual2021},
DPVO \cite{teedDeep2023},
NeRF-VO \cite{naumannNeRFVO2024},
GlORIE-SLAM \cite{zhangGlORIESLAMGloballyOptimized2024}, MonoGS \cite{matsukiGaussian2024}, 
ThermalMonoDepth \cite{shinMaximizingSelfsupervisionThermal2022}, and TOM-GS on the RRXIO dataset. We keep the image resolution at 640$\times$512 for ORB-SLAM3 and LDSO. For other methods, we resize the images to 320$\times$256 to reduce GPU memory usage. NeRF-VO, GlORIE-SLAM, MonoGS, and PhotoSLAM are run in sequential mode. 
We enable global BA for DROID-SLAM and GlORIE-SLAM, and enable loop closure for ORB-SLAM3, LDSO, and Photo-SLAM.

Fig.~\ref{fig:rrxio_odom} reports the ATE using Sim(3) alignment for visual and thermal inputs on RRXIO. As expected, odometry accuracy with coarse thermal images is generally lower than with visual images under well-lit conditions, but surpasses visual performance in challenging settings such as low-light scenes (e.g., mocap\_dark). Geometric methods often fail on thermal sequences. 


For the VIVID dataset, Fig.~\ref{fig:vivid_odom} shows the ATE results for RGB images and for thermal images enhanced with Fieldscale. Learning-based methods significantly outperform geometric methods across most sequences. Importantly, odometry performance using thermal data is generally superior to that using RGB, reflecting the robustness of thermal imaging under variable illumination. Our method achieves leading performance among the learning-based approaches.

\subsection{Visual Rendering Evaluation}

For visual rendering, we compare dense reconstruction methods including NeRF-VO \cite{naumannNeRFVO2024}, GlORIE-SLAM \cite{zhangGlORIESLAMGloballyOptimized2024}, MonoGS \cite{matsukiGaussian2024}, 
PhotoSLAM \cite{huangPhotoSLAMRealtimeSimultaneous2024}, and our TOM-GS. For evaluation, we sample every fifth frame from each sequence, excluding those used for GS or NeRF training. All methods are tested on the RRXIO and VIVID datasets. We report PSNR and LPIPS as in ~\cite{matsukiGaussian2024} to quantify rendering quality. Because the official Photo-SLAM implementation does not compute LPIPS, we report only its PSNR results.

The rendering metrics for the RRXIO sequences are given in Table~\ref{tab:rrxio_render}. GS-based mapping with thermal inputs achieves better rendering quality than with monochrome visual data, especially on dark sequences where thermal imaging is more informative. Our GS-based mapping method attains leading performance on most sequences. However, in large-scale scenes, \eg, campus, our method occasionally fails due to GPU memory exhaustion on a RTX 4090.

The rendering results on the VIVID dataset are reported in Table~\ref{tab:vivid_render}. All thermal rendering results are obtained using the Fieldscale--ZoeDepth configuration. The relatively high PSNR of RGB data on dark sequences, such as indoor\_robust\_dark, is attributable to their limited intensity range. Since most pixels have similarly low intensity values, the mean squared error remains small, making PSNR less discriminative for evaluating rendering quality in these scenes. In contrast, the benefit of thermal imagery is more evident on sequences with varying illumination, such as indoor\_robust\_varying, where TOM-GS achieves higher PSNR using thermal input. These results highlight the advantage of thermal imaging under illumination changes, as RGB appearance varies substantially while thermal observations remain relatively stable.
\begin{figure*}[!t]
\centering

\makebox[\textwidth][c]{%
    \includegraphics[height=1.65cm]{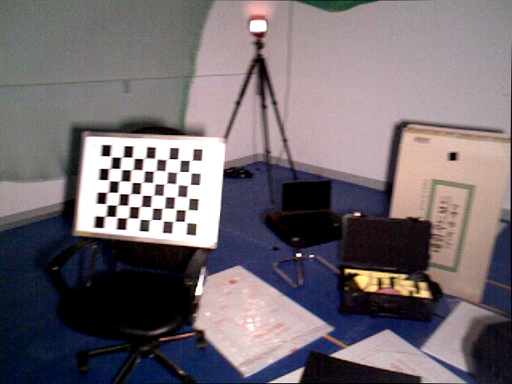}%
    \includegraphics[height=1.65cm]{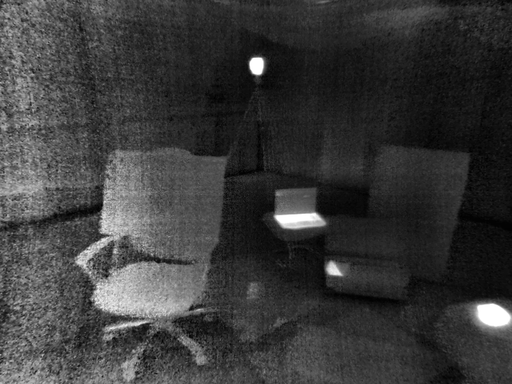}%
    \includegraphics[height=1.65cm]{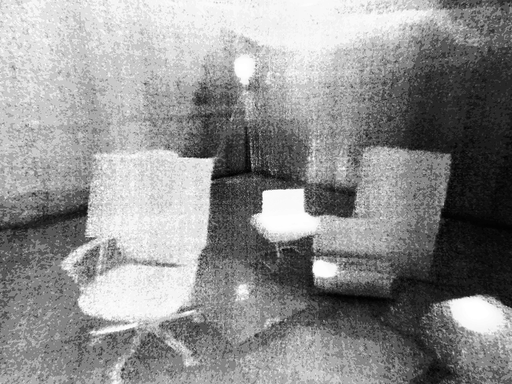}%
    \includegraphics[height=1.65cm]{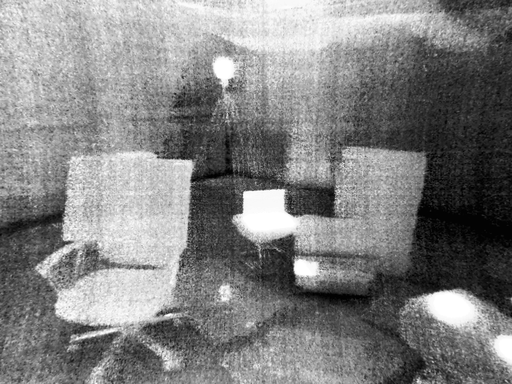}
    \includegraphics[height=1.65cm]{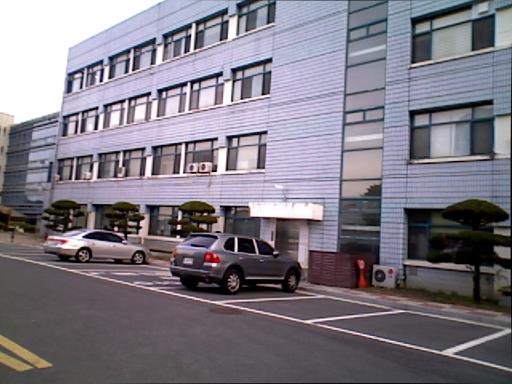}%
    \includegraphics[height=1.65cm]{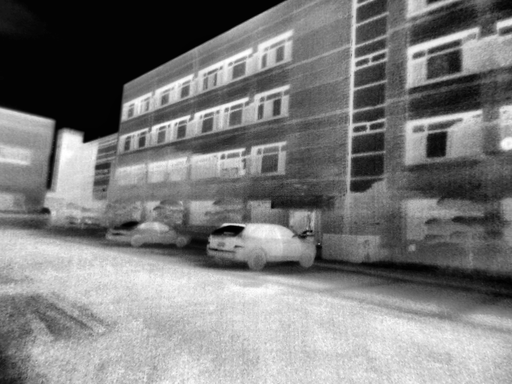}%
    \includegraphics[height=1.65cm]{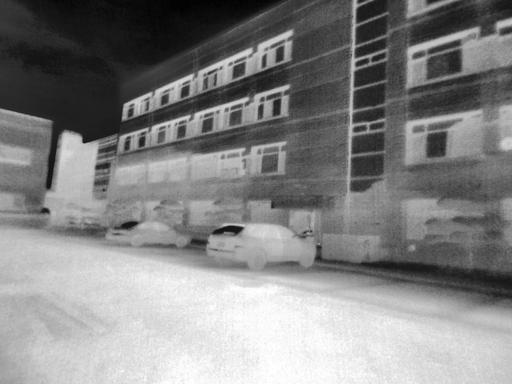}%
    \includegraphics[height=1.65cm]{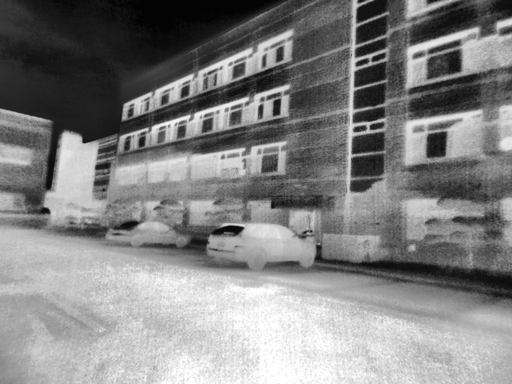}%
}

\makebox[\textwidth][c]{%
    \includegraphics[height=1.65cm]{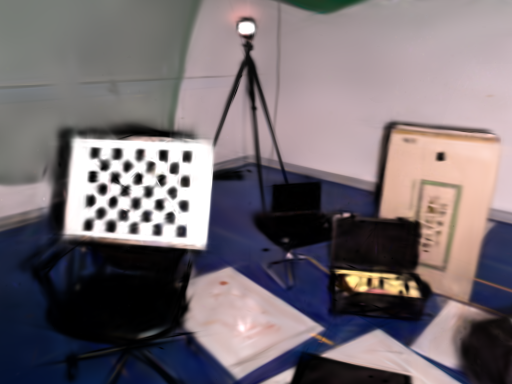}%
    \includegraphics[height=1.65cm]{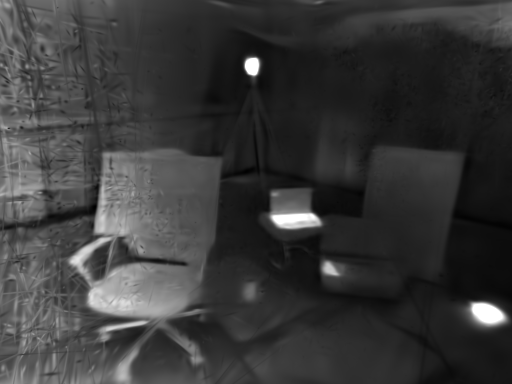}%
    \includegraphics[height=1.65cm]{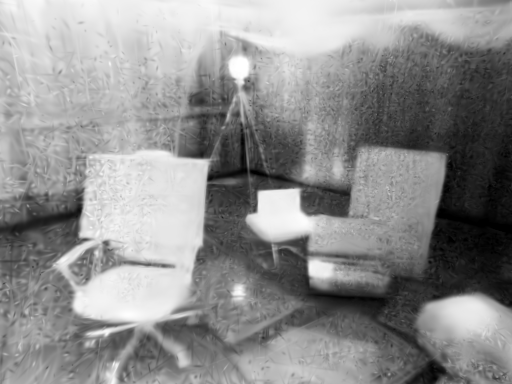}%
    \includegraphics[height=1.65cm]{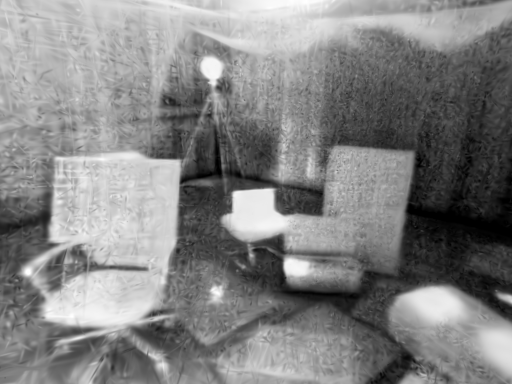}
    \includegraphics[height=1.65cm]{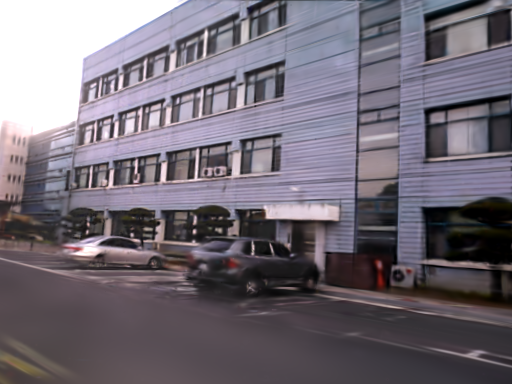}%
    \includegraphics[height=1.65cm]{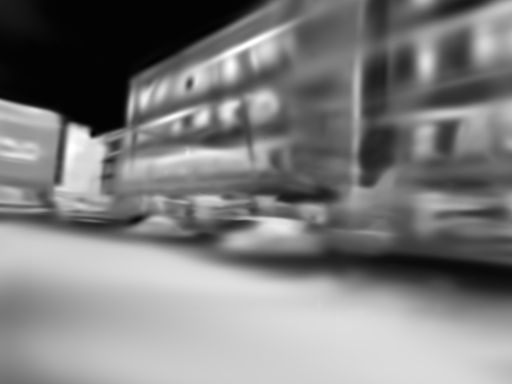}%
    \includegraphics[height=1.65cm]{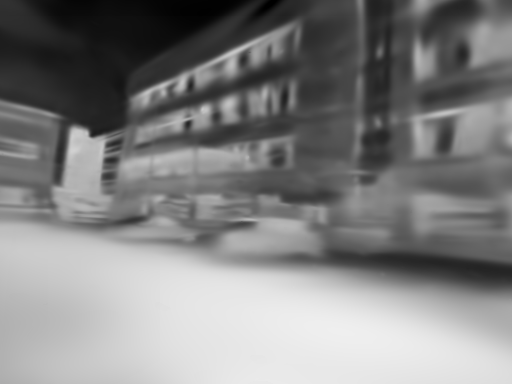}%
    \includegraphics[height=1.65cm]{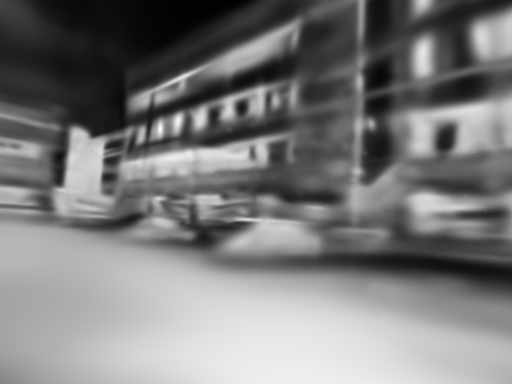}%
}

\makebox[\textwidth][c]{%
    \includegraphics[height=1.65cm]{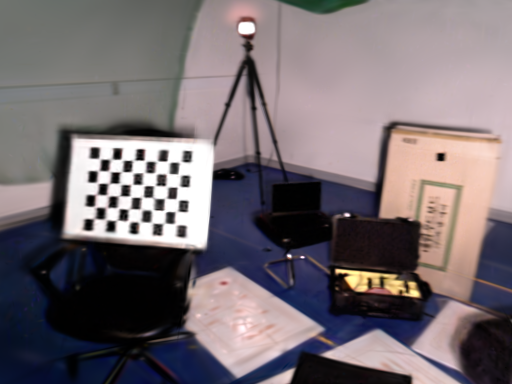}%
    \includegraphics[height=1.65cm]{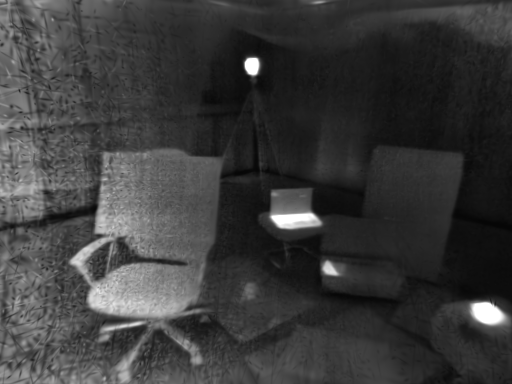}%
    \includegraphics[height=1.65cm]{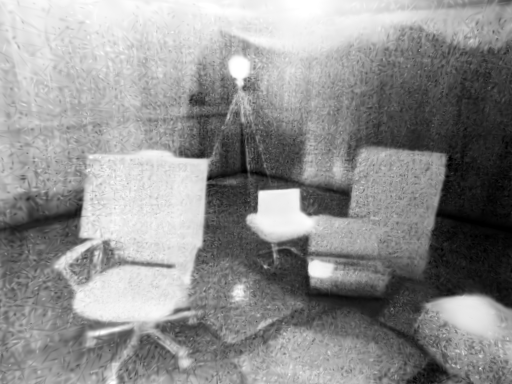}%
    \includegraphics[height=1.65cm]{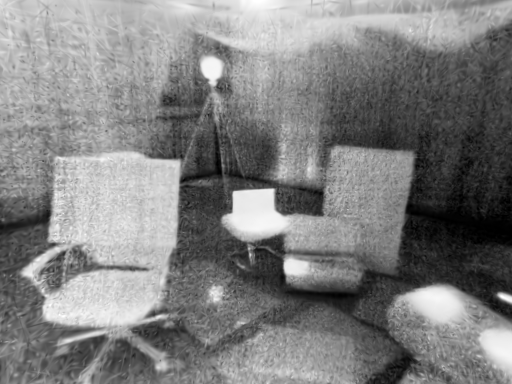}
    \includegraphics[height=1.65cm]{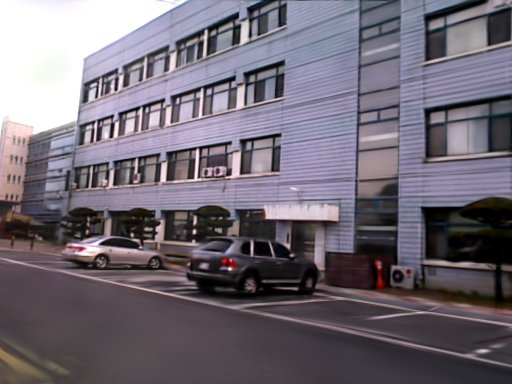}%
    \includegraphics[height=1.65cm]{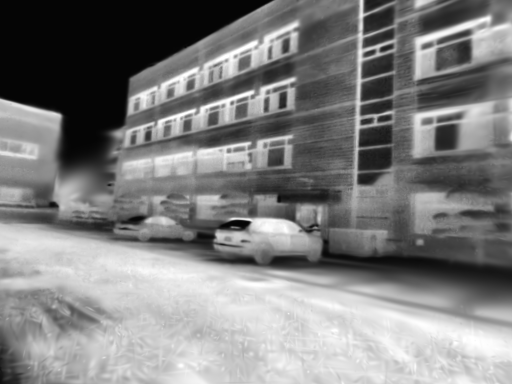}%
    \includegraphics[height=1.65cm]{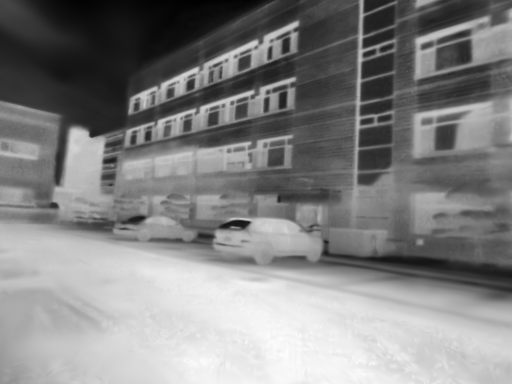}%
    \includegraphics[height=1.65cm]{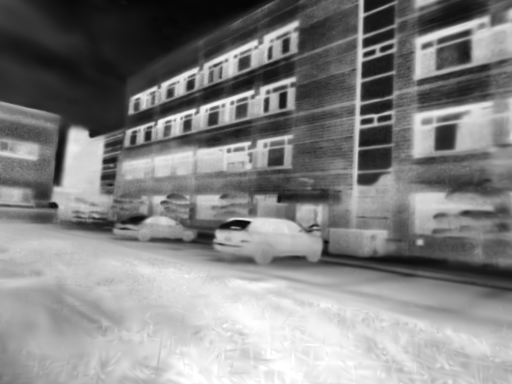}%
}

\vspace{-0.4em}

\makebox[\textwidth][c]{%
    \makebox[0.5\textwidth]{\scriptsize Indoor robust local}%
    \makebox[0.5\textwidth]{\scriptsize Outdoor robust day2}
}

\vspace{0.2em}

\makebox[\textwidth][c]{%
    \includegraphics[height=1.65cm]{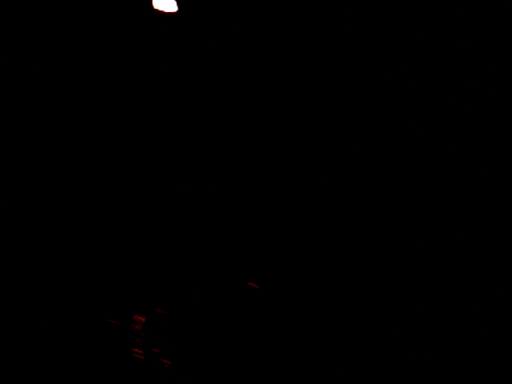}%
    \includegraphics[height=1.65cm]{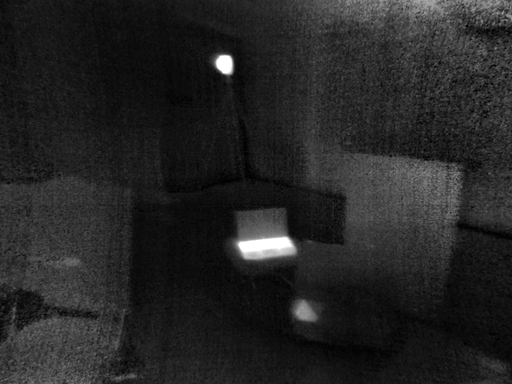}%
    \includegraphics[height=1.65cm]{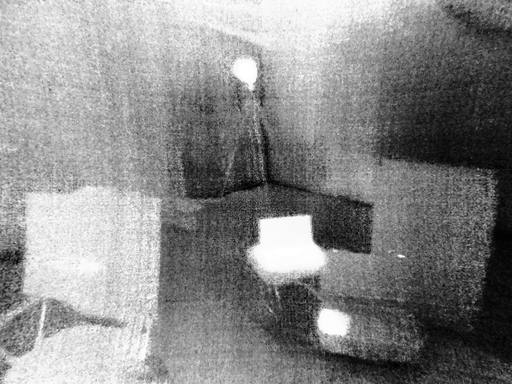}%
    \includegraphics[height=1.65cm]{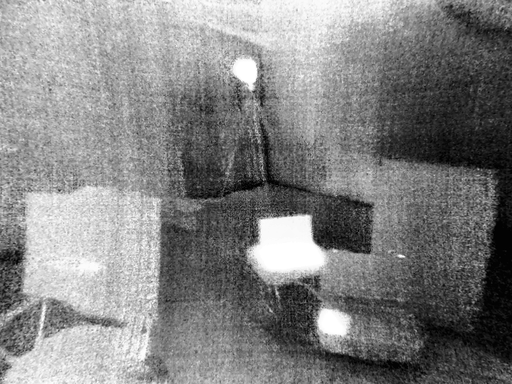}
    \includegraphics[height=1.65cm]{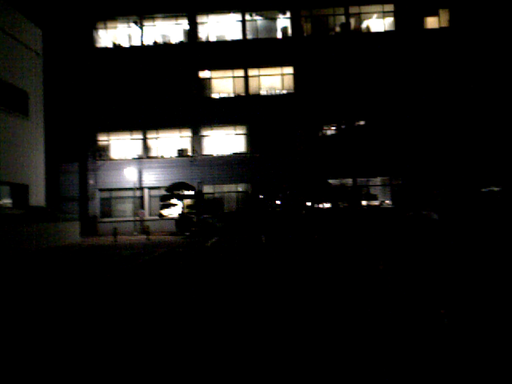}%
    \includegraphics[height=1.65cm]{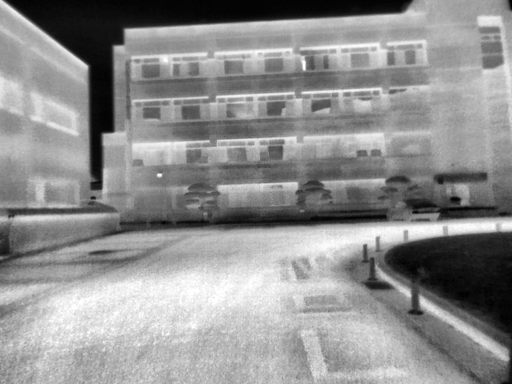}%
    \includegraphics[height=1.65cm]{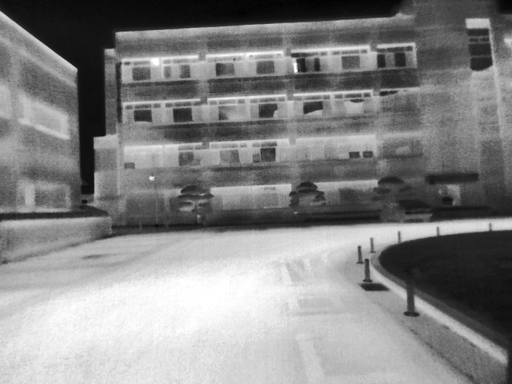}%
    \includegraphics[height=1.65cm]{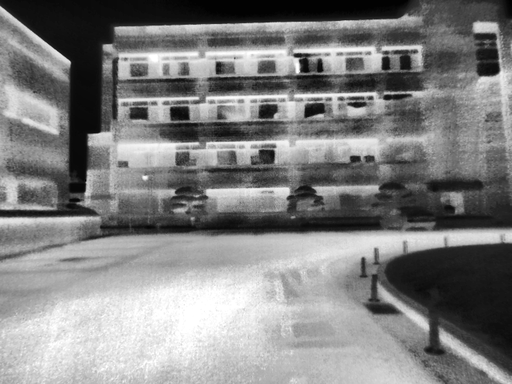}%
}

\makebox[\textwidth][c]{%
    \includegraphics[height=1.65cm]{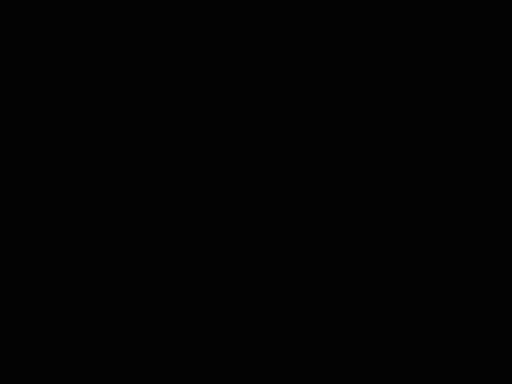}%
    \includegraphics[height=1.65cm]{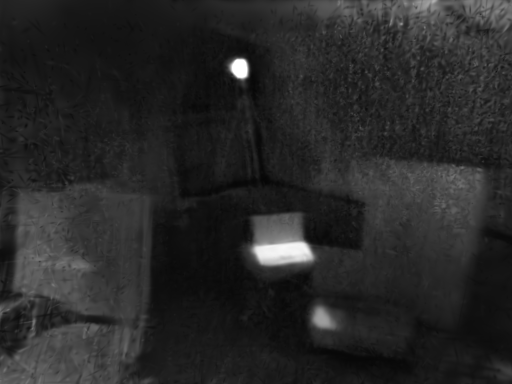}%
    \includegraphics[height=1.65cm]{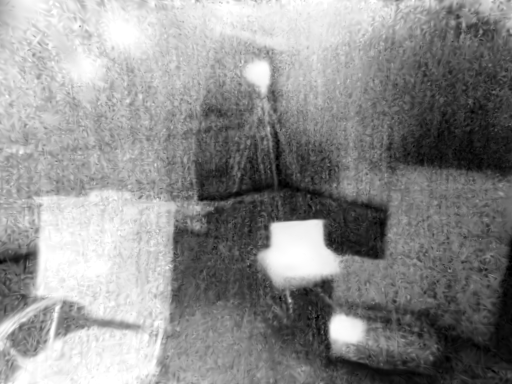}%
    \includegraphics[height=1.65cm]{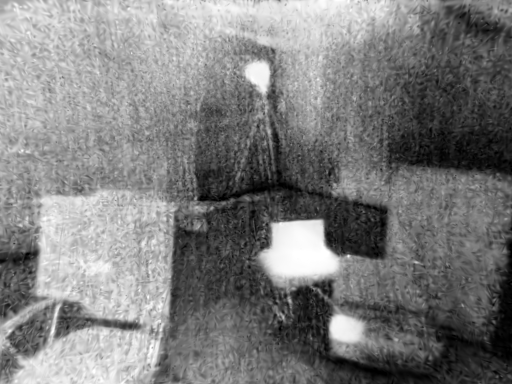}
    \includegraphics[height=1.65cm]{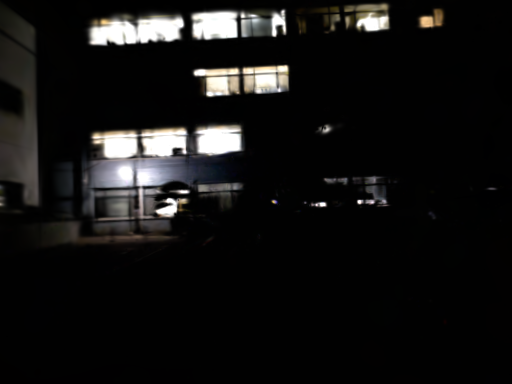}%
    \includegraphics[height=1.65cm]{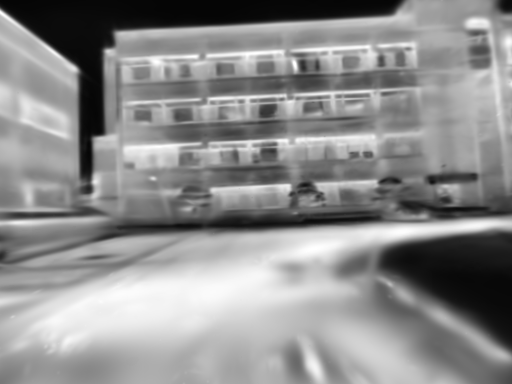}%
    \includegraphics[height=1.65cm]{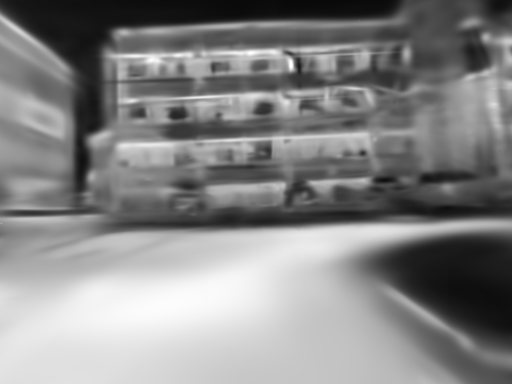}%
    \includegraphics[height=1.65cm]{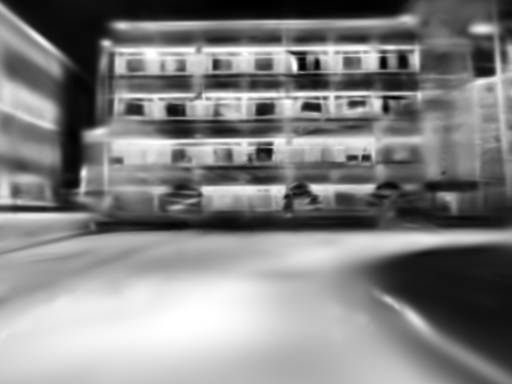}%
}

\makebox[\textwidth][c]{%
    \includegraphics[height=1.65cm]{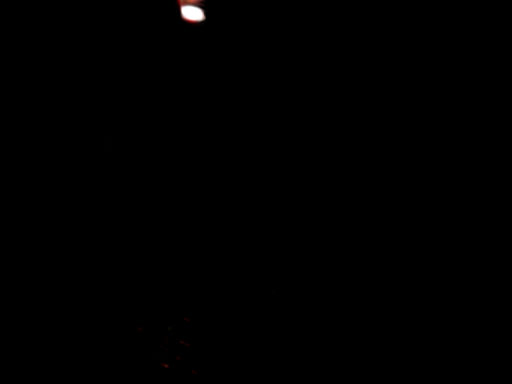}%
    \includegraphics[height=1.65cm]{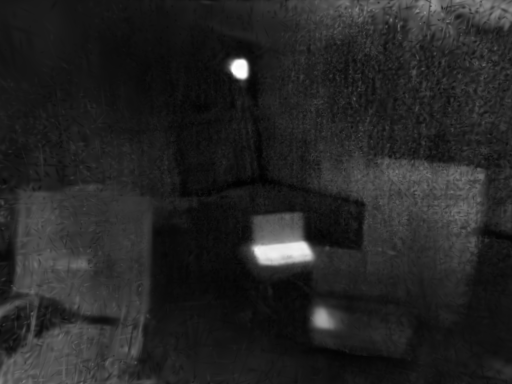}%
    \includegraphics[height=1.65cm]{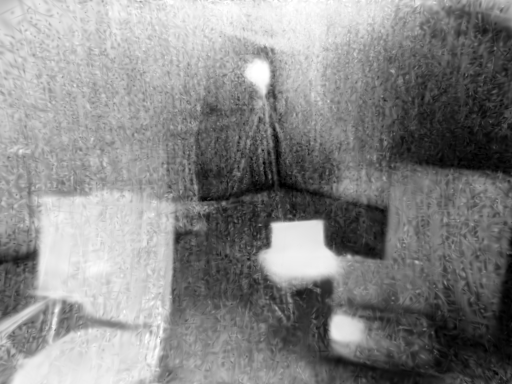}%
    \includegraphics[height=1.65cm]{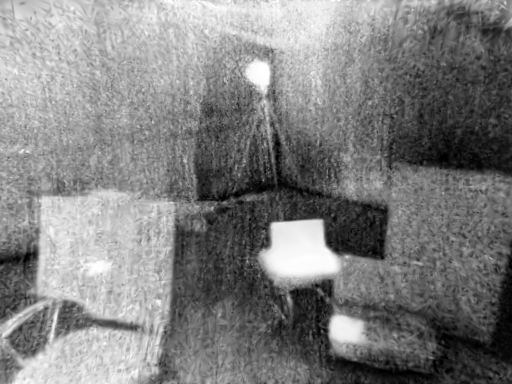}
    \includegraphics[height=1.65cm]{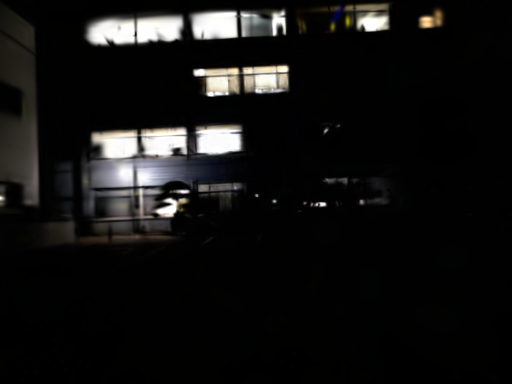}%
    \includegraphics[height=1.65cm]{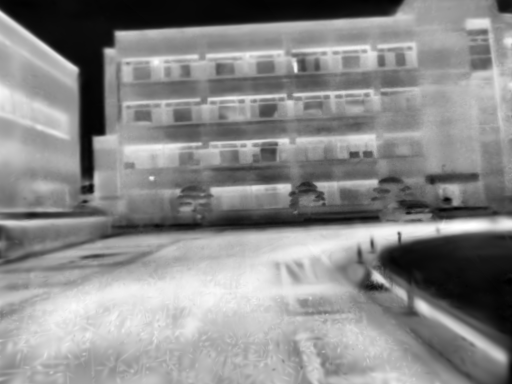}%
    \includegraphics[height=1.65cm]{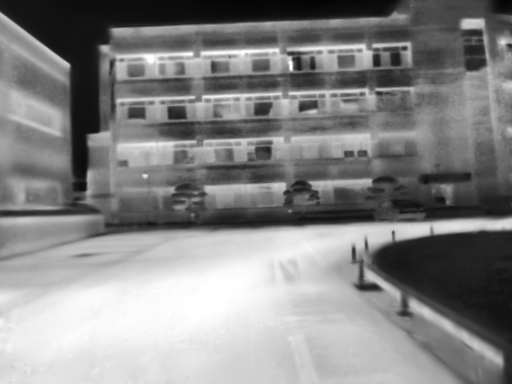}%
    \includegraphics[height=1.65cm]{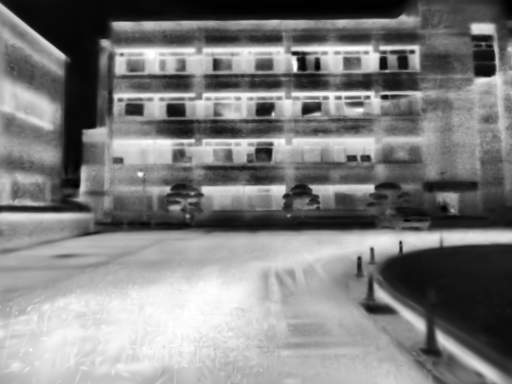}%
}

\vspace{-0.4em}

\makebox[\textwidth][c]{%
    \makebox[0.5\textwidth]{\scriptsize Indoor unstable dark}%
    \makebox[0.5\textwidth]{\scriptsize Outdoor robust night1}
}

\vspace{0.2em}

\caption{Rendering results for sampled RGB and thermal images from four VIVID sequences:
indoor robust local, outdoor robust day2, indoor unstable dark, outdoor robust night1.
In each group, columns from left to right show RGB, Thermal-Fieldscale, Thermal-naive, and Thermal-Shin results; rows from top to bottom show reference images, rendered images without a depth prior, and rendered images with the ZoeDepth prior.
}
\label{fig:comparison}
\end{figure*} 

\begin{table*}[t]
\centering
\tiny

\caption{Rendering results of TOM-GS on the VIVID sequences converted to 8-bit using different enhancement methods, including Fieldscale (fs), Naïve, and Shin's method, both without depth priors and with depth priors estimated by Depth Anything V2, Lotus, Metric3D, and ZoeDepth. Best average values are \textbf{bolded}, and second-best values are \underline{underlined}.}
\label{tab:ablation}

\resizebox{\textwidth}{!}{
\begin{tabular}{c|c|c|c|c|c|c|c|c|c|c|c|c|c|c|c|c}
\Xhline{1.5pt}
Method & Metric &
IAD & IAG & IAL & IRD & IRG & IRL & IRV &
IUD & IUG & IUL & ORD1 & ORD2 & ORN1 & ORN2 & Avg. \\
\Xhline{1.5pt}

\multirow{2}{*}{fieldscale}
& PSNR$\uparrow$ & 27.82 & 22.12 & 28.86 & 27.29 & 27.50 & 27.52 & 24.23 & 28.49 & 26.53 & 29.80 & 22.27 & 20.57 & 22.88 & 20.92 & 25.49 \\
& LPIPS$\downarrow$ & 0.31 & 0.52 & 0.32 & 0.35 & 0.37 & 0.37 & 0.53 & 0.28 & 0.28 & 0.28 & 0.38 & 0.39 & 0.37 & 0.40 & 0.37 \\
\hline

\multirow{2}{*}{naive}
& PSNR$\uparrow$ & 22.34 & 24.51 & 19.07 & 20.58 & 27.35 & 21.32 & 20.34 & 21.51 & 27.34 & 21.27 & 28.60 & 27.80 & 28.33 & 25.08 & 23.96 \\
& LPIPS$\downarrow$ & 0.33 & 0.42 & 0.44 & 0.37 & 0.38 & 0.39 & 0.37 & 0.34 & 0.29 & 0.35 & 0.26 & 0.27 & 0.27 & 0.33 & 0.35 \\
\hline

\multirow{2}{*}{shin}
& PSNR$\uparrow$ & 21.32 & 23.02 & 20.94 & 21.22 & 25.25 & 21.85 & 20.96 & 21.58 & 25.12 & 21.38 & 26.00 & 24.56 & 25.49 & 24.90 & 23.11 \\
& LPIPS$\downarrow$ & 0.34 & 0.38 & 0.34 & 0.33 & 0.31 & 0.35 & 0.33 & 0.30 & 0.28 & 0.31 & 0.27 & 0.33 & 0.30 & 0.31 & 0.32 \\
\hline

\multirow{2}{*}{Depthany-fs}
& PSNR$\uparrow$ & 24.06 & 26.27 & 25.03 & 27.97 & 28.22 & 29.59 & 29.30 & 26.80 & 27.17 & 30.15 & 14.11 & 16.81 & 27.69 & 24.15 & \underline{25.52} \\
& LPIPS$\downarrow$ & 0.45 & 0.37 & 0.44 & 0.28 & 0.27 & 0.23 & 0.22 & 0.27 & 0.28 & 0.21 & 0.56 & 0.43 & 0.23 & 0.27 & 0.32 \\
\hline

\multirow{2}{*}{Depthany-naive}
& PSNR$\uparrow$ & 20.67 & 25.89 & 22.39 & 20.42 & 28.19 & 22.06 & 20.86 & 21.54 & 28.21 & 21.31 & 27.96 & 29.15 & 29.89 & 28.35 & 24.78 \\
& LPIPS$\downarrow$ & 0.50 & 0.40 & 0.42 & 0.35 & 0.26 & 0.30 & 0.31 & 0.38 & 0.29 & 0.31 & 0.26 & 0.23 & 0.23 & 0.23 & 0.32 \\
\hline

\multirow{2}{*}{Depthany-shin}
& PSNR$\uparrow$ & 21.88 & 25.35 & 19.82 & 21.62 & 25.02 & 22.48 & 21.20 & 21.78 & 25.25 & 21.64 & 26.85 & 26.67 & 27.58 & 26.08 & 23.80 \\
& LPIPS$\downarrow$ & 0.35 & 0.37 & 0.46 & 0.29 & 0.30 & 0.27 & 0.28 & 0.29 & 0.36 & 0.28 & 0.25 & 0.25 & 0.24 & 0.25 & \textbf{0.30} \\
\hline

\multirow{2}{*}{Lotus-fs}
& PSNR$\uparrow$ & 27.68 & 21.42 & 28.70 & 26.09 & 24.45 & 24.57 & 25.78 & 25.58 & 26.40 & 29.84 & 23.46 & 20.55 & 21.59 & 20.86 & 24.78 \\
& LPIPS$\downarrow$ & 0.28 & 0.58 & 0.33 & 0.37 & 0.49 & 0.49 & 0.45 & 0.47 & 0.28 & 0.27 & 0.33 & 0.39 & 0.40 & 0.41 & 0.39 \\
\hline

\multirow{2}{*}{Lotus-naive}
& PSNR$\uparrow$ & 22.11 & 24.60 & 16.77 & 19.81 & 27.13 & 21.23 & 20.33 & 21.13 & 27.32 & 21.49 & 28.14 & 28.05 & 28.01 & 25.45 & 23.68 \\
& LPIPS$\downarrow$ & 0.33 & 0.35 & 0.63 & 0.44 & 0.37 & 0.39 & 0.37 & 0.38 & 0.30 & 0.34 & 0.27 & 0.28 & 0.28 & 0.32 & 0.36 \\
\hline

\multirow{2}{*}{Lotus-shin}
& PSNR$\uparrow$ & 21.39 & 22.54 & 21.00 & 20.78 & 24.66 & 21.82 & 20.56 & 21.46 & 25.24 & 21.58 & 25.54 & 24.63 & 25.61 & 24.26 & 22.93 \\
& LPIPS$\downarrow$ & 0.32 & 0.38 & 0.34 & 0.35 & 0.38 & 0.35 & 0.34 & 0.31 & 0.27 & 0.31 & 0.28 & 0.31 & 0.31 & 0.33 & 0.33 \\
\hline

\multirow{2}{*}{Metric3D-fs}
& PSNR$\uparrow$ & 26.75 & 20.91 & 29.84 & 26.38 & 22.97 & 27.96 & 27.49 & 28.31 & 26.64 & 30.18 & 22.30 & 20.94 & 23.32 & 23.02 & 25.50 \\
& LPIPS$\downarrow$ & 0.29 & 0.59 & 0.32 & 0.37 & 0.56 & 0.34 & 0.32 & 0.27 & 0.29 & 0.27 & 0.33 & 0.39 & 0.31 & 0.34 & 0.36 \\
\hline

\multirow{2}{*}{Metric3D-naive}
& PSNR$\uparrow$ & 22.29 & 22.83 & 15.87 & 19.47 & 27.30 & 21.19 & 20.23 & 21.24 & 27.53 & 21.30 & 28.19 & 27.81 & 28.10 & 25.64 & 23.50 \\
& LPIPS$\downarrow$ & 0.33 & 0.43 & 0.70 & 0.45 & 0.38 & 0.39 & 0.38 & 0.38 & 0.28 & 0.33 & 0.26 & 0.27 & 0.28 & 0.32 & 0.37 \\
\hline

\multirow{2}{*}{Metric3D-shin}
& PSNR$\uparrow$ & 21.34 & 22.58 & 19.92 & 20.73 & 24.28 & 21.70 & 20.71 & 21.50 & 25.28 & 21.70 & 25.62 & 24.75 & 25.44 & 24.86 & 22.89 \\
& LPIPS$\downarrow$ & 0.33 & 0.40 & 0.47 & 0.36 & 0.40 & 0.35 & 0.34 & 0.31 & 0.28 & 0.31 & 0.29 & 0.30 & 0.31 & 0.31 & 0.34 \\
\hline

\multirow{2}{*}{ZoeDepth-fs}
& PSNR$\uparrow$ & 25.79 & 26.97 & 28.44 & 27.95 & 27.16 & 29.91 & 29.21 & 28.45 & 25.72 & 30.27 & 14.35 & 21.87 & 24.94 & 25.97 & \textbf{26.22} \\
& LPIPS$\downarrow$ & 0.39 & 0.30 & 0.37 & 0.27 & 0.29 & 0.23 & 0.22 & 0.25 & 0.36 & 0.21 & 0.56 & 0.29 & 0.26 & 0.24 & \textbf{0.30} \\
\hline

\multirow{2}{*}{ZoeDepth-naive}
& PSNR$\uparrow$ & 20.72 & 27.37 & 21.02 & 20.51 & 28.30 & 21.76 & 20.70 & 21.96 & 28.21 & 21.31 & 27.87 & 28.43 & 29.88 & 26.88 & 24.64 \\
& LPIPS$\downarrow$ & 0.46 & 0.28 & 0.38 & 0.34 & 0.25 & 0.30 & 0.31 & 0.33 & 0.29 & 0.30 & 0.26 & 0.24 & 0.24 & 0.25 & \textbf{0.30} \\
\hline

\multirow{2}{*}{ZoeDepth-shin}
& PSNR$\uparrow$ & 21.71 & 24.09 & 19.42 & 21.67 & 24.86 & 22.34 & 21.28 & 21.99 & 24.68 & 21.76 & 26.87 & 26.71 & 27.49 & 26.18 & 23.65 \\
& LPIPS$\downarrow$ & 0.36 & 0.43 & 0.47 & 0.30 & 0.29 & 0.28 & 0.28 & 0.29 & 0.36 & 0.28 & 0.25 & 0.24 & 0.25 & 0.25 & \underline{0.31} \\
\Xhline{1.5pt}

\end{tabular}
}
\end{table*}

\subsection{Qualitative Evaluation}

We further qualitatively evaluate the reconstruction quality of TOM-GS on the VIVID dataset. Figure~\ref{fig:comparison} presents representative rendering
results using RGB images and thermal images processed with three different enhancement methods. Benefiting from the proposed depth prior integration, TOM-GS preserves richer scene details and produces sharper and more faithful reconstructions, particularly in scenes with large depth ranges, where distant structures exhibit limited spatial detail and weak multi-view parallax.


\subsection{Ablation Study}
\label{subsec:ablation}


We conduct ablation studies on the VIVID dataset to analyze the effects of monocular depth priors and thermal image enhancement methods. The evaluated depth models include
Depth Anything V2 \cite{yangDepthAnythingV22024}, Lotus \cite{heLotus2025},
Metric3D \cite{yinMetric3DZeroshotMetric2023}, and ZoeDepth \cite{bhatZoeDepthZeroshotTransfer2023}, while Fieldscale \cite{gilFieldscale2024}, 
Shin's method \cite{shinMaximizingSelfsupervisionThermal2022}, and a naive CLAHE-based method \cite{bradskiLearningOpenCVComputer2008} are compared for thermal image enhancement.

As shown in Table~\ref{tab:ablation}, Fieldscale generally provides the most effective thermal enhancement, 
and the Fieldscale–ZoeDepth combination achieves the best overall performance. Compared with the configuration without a depth prior, it increases the average PSNR from 25.49\,dB to 26.22\,dB and reduces LPIPS from 0.37 to 0.30. The monocular prior noticeably improves the stability of Gaussian mapping. Without the prior, excessive Gaussian pruning is observed during later mapping stages, suggesting less stable geometric initialization and optimization. By providing complementary geometric constraints, the prior stabilizes Gaussian optimization and improves rendering quality.

However, not all depth priors consistently improve the results, as Lotus and Metric3D degrade rendering quality in some configurations. This suggests that the effectiveness of a depth prior depends primarily on its compatibility with thermal imagery and the adopted preprocessing method, rather than on model size alone.

\subsection{GPU Memory Consumption}
TOM-GS runs online on an NVIDIA RTX 4090 GPU. Its GPU memory consumption mainly consists of three components: the learning-based odometry frontend, the Gaussian mapping backend, and the monocular depth prior network. The odometry module requires approximately 6.7\,GB of GPU memory, while the Gaussian mapping module typically consumes 2--8\,GB depending on the number of Gaussians. The memory overhead of the monocular depth prior depends on the adopted foundation model. 
Overall, peak GPU memory usage during online operation ranges from 11 to 20 GB. When a large depth model is used, the monocular depth network accounts for the largest share of memory consumption.

\section{CONCLUSIONS}
We presented TOM-GS, a monocular thermal odometry and mapping system that integrates learning-based dense tracking with depth-guided 3D Gaussian Splatting. By leveraging monocular depth priors to stabilize pose estimation and supervise Gaussian geometry, the proposed framework enables consistent and dense reconstruction from low-texture thermal imagery.
Experiments on the RRXIO and VIVID benchmarks demonstrate that TOM-GS outperforms classical geometric thermal SLAM in tracking accuracy and robustness, while thermal sensing provides improved resilience under degraded illumination compared to RGB. The results highlight the potential of learning-based dense SLAM for autonomy in adverse environments. Future work will focus on improving scalability through submap-based strategies for large-scale deployment.


\bibliographystyle{IEEEtran}
\bibliography{IEEEabrv,zotero}

\end{document}